\documentclass[11pt]{article}

\usepackage[final]{acl}

\usepackage{times}
\usepackage{latexsym}
\usepackage{amssymb}
\usepackage[T1]{fontenc}
\usepackage[utf8]{inputenc}
\usepackage{microtype}
\usepackage{booktabs}
\usepackage{multirow}
\usepackage{graphicx}
\usepackage{xcolor}
\usepackage{colortbl}
\usepackage{paralist}
\usepackage{tabularx}
\usepackage{textcomp}
\usepackage{fvextra}
\usepackage{tikz}
\usetikzlibrary{arrows.meta}

\usepackage{annotates}
\providecommentcommand{rom}{blue}{Roman}
\providecommentcommand{todo}{red}{ToDo}

\definecolor{expcol}{HTML}{CDE3F5}
\definecolor{stimcol}{HTML}{F9D9B5}
\definecolor{cuecol}{HTML}{DCEFD0}

\newcommand{\prelim}[1]{#1}
\newcommand{\hub}{BamNLP Emotion Hub}
\newcommand{\res}[1]{\textsc{#1}}

\title{\raisebox{-0.3\height}{\includegraphics[height=1.6em]{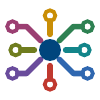}}\hspace{0.4em}%
  The BamNLP Emotion Hub:\\ Towards Generalizing Models for
  Multiple Tasks in Emotion Analysis%
}

\author{Roman Klinger \\
  Fundamentals of Natural Language Processing, University of Bamberg, Germany \\
  \texttt{roman.klinger@uni-bamberg.de}}

\begin{document}
\maketitle

\begin{abstract}
  Emotion analysis in text comprises several tasks, from the
  classification of emotion categories over valence and arousal
  regression to appraisal prediction, emotion component detection, and
  the labeling of experiencers and stimuli. Each of these tasks has
  been studied on corpora with their own label sets, annotation
  perspectives, and kinds of text. Models trained on one corpus are
  rarely evaluated on another. Therefore, practitioners who want to
  analyse their own texts do not know which model to trust, and
  researchers do not know which phenomena remain unsolved. To make the
  current situation more transparent, we present the \hub{}, a website
  that brings together \prelim{48} emotion corpora (\prelim{24} of our
  own and \prelim{24} of other groups) in one format, models trained
  on each corpus and on combinations of corpora, and the evaluation of
  every model on every corpus. The cross-corpus evaluation shows that
  models reach a median of \prelim{65\%} of the performance of the
  model trained on the target corpus when both share the kind of text,
  but only \prelim{53\%} across kinds of text; valence and arousal
  transfer well (\prelim{86--90\%}), appraisals less so
  (\prelim{61--64\%}). We further contribute a general model, trained
  jointly on all human-annotated corpora, that predicts categories,
  valence--arousal, appraisals and emotion components, in two
  variants: a perspective model, which is told whose emotion is meant,
  and a no-perspective model, which we use in the case studies. It
  comes close to the single-corpus models on their own data but drops
  on texts that no corpus covers. We illustrate the use of the hub
  with various case studies.
\end{abstract}

\section{Introduction}
\label{sec:intro}

\begin{figure}[t]
  \centering\small
  \begin{tikzpicture}[
      span/.style={rounded corners=1pt, inner sep=1.5pt}]
    \node[text width=\columnwidth-14pt, align=left, font=\small] (s) at (0,0) {%
      ``\tikz[baseline]\node[span, fill=expcol, anchor=base]{Maria};
      smiled with
      \tikz[baseline]\node[span, fill=cuecol, anchor=base]{joy};
      when
      \tikz[baseline]\node[span, fill=stimcol, anchor=base]{her daughter came home for Christmas};.''};
  \end{tikzpicture}\\[3pt]
  {\scriptsize\renewcommand{\arraystretch}{1.15}%
  \begin{tabularx}{\columnwidth}{@{}>{\sffamily\raggedleft\arraybackslash}p{1.9cm}X@{}}
    Roles & \colorbox{expcol}{experiencer} \colorbox{stimcol}{stimulus/target} \colorbox{cuecol}{cue} \\
    Categories & joy .98 (next: no emotion .05) \\
    Valence/arousal & valence .87, arousal .30 \\
    Event gate & describes an event ($p>.99$) $\Rightarrow$ appraisals shown \\
    Appraisals & certain .97, pleasant .95, attention .93, goal-conducive .77, other responsible .65;
      effort .02, unpleasant .08 \\
    Components & cognitive appraisal .73, subjective feeling .61, motor expression .58 \\
  \end{tabularx}}
  \caption{One text with the output layers of the \hub{}'s general model and role tagger (actual outputs on an
    invented sentence): emotion roles, emotion categories, valence and arousal, the event gate, appraisals, and
    emotion components. Appraisals show the five highest and two lowest of 16 values.}
  \label{fig:teaser}
\end{figure}
 
Emotion analysis is the task of inferring emotions from text. It
comprises a set of related tasks
\citep{klinger-2023-event,plaza-del-arco-etal-2024-emotion}: the
classification of a text into emotion categories such as anger, fear
or joy
\citep{alm-etal-2005-emotions,mohammad-2012-emotional,demszky-etal-2020-goemotions};
the regression of affective dimensions such as valence and arousal
\citep{buechel-hahn-2017-emobank,preotiuc-pietro-etal-2016-modelling}
or intensity \citep{mohammad-bravo-marquez-2017-wassa}; the prediction
of cognitive appraisals, i.e., how a person evaluates an event
\citep{hofmann-etal-2020-appraisal,troiano-etal-2023-dimensional}; the
detection of emotion components after Scherer's component process
model \citep{casel-etal-2021-emotion}; and the labeling of the
semantic roles of an emotion: who feels it (experiencer), what caused
it (stimulus), and towards whom it is directed (target)
\citep{mohammad-etal-2014-semantic,kim-klinger-2018-feels,bostan-etal-2020-goodnewseveryone,campagnano-etal-2022-srl4e}.
All of these tasks additionally depend on the perspective: a text may
express the writer's own emotion, describe the emotion of a person in
the text, or evoke an emotion in a reader
\citep{buechel-hahn-2017-emobank,troiano-etal-2019-crowdsourcing}.
Figure~\ref{fig:teaser} shows one text with annotations.

\begin{table*}[t]
  \centering\footnotesize
  \setlength{\tabcolsep}{2pt}
  \begin{tabular}{@{}lrlcccccccc@{}}
    \toprule
    & & & \multicolumn{5}{c}{Tasks} & & & \\
    \cmidrule(lr){4-8}
    Effort & Corpora & Languages & Cat. & VA & App. & Comp. & Roles & Persp. & Models & X-corpus \\
    \cmidrule(r){1-1}\cmidrule(rl){2-2}\cmidrule(rl){3-3}\cmidrule(l){4-8}\cmidrule(l){9-11}
    \res{UnifiedEmotions} \citep{bostan-klinger-2018-analysis} & 14 & en & \checkmark & & & & & & & \checkmark \\
    \res{Srl4e} \citep{campagnano-etal-2022-srl4e} & 6 & en, zh & \checkmark & & & & \checkmark & & & \checkmark \\
    \citet{el-haj-takanami-2023-unifying} & 5 & en & \checkmark & \checkmark & & & & & & \\
    \citet{koufakou-etal-2024-towards} & 11 & en & \checkmark & & & & & & & \checkmark \\
    \res{TweetEval} \citep{barbieri-etal-2020-tweeteval} & 1 & en & \checkmark & & & & & & \checkmark & \\
    \res{Brighter} \citep{muhammad-etal-2025-brighter} & 1 (new) & 28 & \checkmark & & & & & & & \\
    \res{EmoBench} \citep{sabour-etal-2024-emobench} & 1 (new) & en, zh & \checkmark & & & & & & & \\
    \res{TweetNLP} \citep{camacho-collados-etal-2022-tweetnlp} & 1 & en & \checkmark & & & & & & \checkmark & \\
    \res{EmoLLMs} \citep{liu-etal-2024-emollms} & 1 (+9) & en & \checkmark & \checkmark & & & & & \checkmark & \checkmark \\
    \midrule
    \hub{} & \prelim{48} & en, de & \checkmark & \checkmark & \checkmark & \checkmark & \checkmark & \checkmark & \checkmark & \checkmark \\
    \bottomrule
  \end{tabular}
  \caption{Efforts that bring emotion corpora or models together. ``new'': one newly annotated data set; ``+9'': data
    sets used for evaluation only. Cat.: categories, VA: valence--arousal or intensity, App.: appraisals, Comp.: emotion
    components, Persp.: annotation perspectives distinguished, X-corpus: cross-corpus evaluation. \res{TweetEval} and
    \res{TweetNLP} cover emotions as one of several tasks.}
  \label{tab:unification}
\end{table*}

Surveys document the methods and resources of emotion analysis, recently also with large language models (LLMs;
Section~\ref{sec:surveys}). Corpora have been brought together before, but each effort covers one task, such as emotion
categories, emotion roles or affective dimensions (Section~\ref{sec:unification}). Models are shared in repositories such
as the Hugging Face Hub, typically trained on a few corpora of one kind of text and with one label set.

Despite these efforts, a user who wants to analyse emotions in their own texts faces three limitations.
\begin{inparaenum}[(1)]
\item Existing unifications cover one task (categories, roles, or dimensions) and do not offer
  models for all of them.
\item Models are rarely evaluated outside the corpus they were trained on, so it is unknown how well a model
  works on a new kind of text, for a new perspective, or in another language.
\item Appraisals, emotion components, and roles have been studied on only a few corpora.
\end{inparaenum}

The \hub{} addresses these limitations or makes them transparent. It
is a
website\footnote{\url{https://www.emotionanalysis.de}}
that (a) lists
and describes emotion corpora, ours and those of other groups, in one
format with the original labels kept; (b) offers models trained on
each corpus and general models trained on many of them, for
categories, valence--arousal, appraisals, and emotion components, and
role labeling models; (c) reports the evaluation of every model on
every corpus; and (d) helps users choose a corpus or model. We pursue
the following questions with the hub: \textbf{RQ1}: How well do
emotion analysis models generalise across corpora, kinds of text,
perspectives and languages?  \textbf{RQ2}: Can a single model predict
all emotion representations across corpora as well as specialised
models?  \textbf{RQ3}: Which questions from the social sciences and
humanities can such models answer at scale?  Our contributions are the
hub itself (Section~\ref{sec:hub}), findings from the cross-corpus
evaluation (Section~\ref{sec:findings}), and case studies
(Section~\ref{sec:casestudies}).\footnote{We see the website as a
  resource that receives updates over time (new corpora, models and
  label mappings); these updates will influence the experimental
  results.}

\section{Related Work}
\label{sec:related}

\subsection{Surveys on Emotion Analysis}
\label{sec:surveys}
The field is well documented, with reviews of methods
\citep{canales-martinez-barco-2014-emotion,mohammad-2016-sentiment},
deep learning
\citep{acheampong-etal-2020-text,alswaidan-menai-2020-survey,deng-ren-2023-survey},
conversations \citep{wu-etal-2025-multimodal}, ethics and best
practices \citep{mohammad-2022-ethics-sheet,mohammad-2023-best},
literary studies \citep{kim-klinger-2019-survey}, and roles and
appraisals \citep{klinger-2023-event,stajner-klinger-2023-emotion}.
\citet{plaza-del-arco-etal-2024-emotion} review 154 publications and
find that emotion analysis lacks a standardised terminology, that the
categories of the two dominant theories fit the task poorly, and that
demographic and cultural factors are largely ignored. Recent reviews
cover LLMs
\citep{zhang-etal-2024-affective,song-etal-2025-large,hegde-jayalath-2025-emotions},
for which emotion analysis has become a common evaluation task
\citep{zhang-etal-2024-sentiment,sabour-etal-2024-emobench,belay-etal-2025-evaluating}.

\subsection{Related Unification Efforts}
\label{sec:unification}

Table~\ref{tab:unification} compares the \hub{} with other efforts
that bring emotion corpora or models together.  \res{UnifiedEmotions}
\citep{bostan-klinger-2018-analysis} converted 14 corpora into one
format, mapped their labels to a shared set, and trained one model per
corpus to test it on every other corpus; the \hub{} continues this
work with more corpora, more tasks, and transformer
models. \res{Srl4e} \citep{campagnano-etal-2022-srl4e} unifies six
corpora with emotion role annotations (cue, experiencer, target,
stimulus) in English and Chinese and evaluates role labeling across
them.  \citet{el-haj-takanami-2023-unifying} map the categories of
five corpora to valence, arousal and dominance and train one model on
the combined data; \citet{koufakou-etal-2024-towards} combine eleven
English single-label corpora and compare training on the combined data
with training on each corpus. Benchmarks such as \res{TweetEval} and
\res{SuperTweetEval}
\citep{barbieri-etal-2020-tweeteval,antypas-etal-2023-supertweeteval}
and toolkits such as \res{TweetNLP} include one emotion data set among
other tasks; \res{Brighter} contributes new annotations in 28
languages and \res{EmoBench} new annotations for LLM evaluation,
instead of unifying existing ones; \res{EmoLLMs}
\citep{liu-etal-2024-emollms} fine-tune LLMs on instruction data from
the five subtasks of SemEval-2018 Task~1 (emotion intensity, valence,
multi-label emotions) and evaluate them on further data sets.

Models are shared in repositories such as the Hugging Face Hub, where
emotion classifiers \citep[e.g.,][]{hartmann-2022-emotion} and
toolkits
\citep{perez-etal-2021-pysentimiento,camacho-collados-etal-2022-tweetnlp}
are available.  None of these efforts covers appraisals or emotion
components, distinguishes annotation perspectives, or offers models
for all tasks together with their cross-corpus evaluation.

\section{The \hub{}}
\label{sec:hub}

\paragraph{Corpora.}
The hub contains \prelim{24} corpora created by our group and \prelim{24} corpora of other groups (Appendix~\ref{app:corpora}, Table~\ref{tab:corpora}). Every corpus is converted
into one JSON Lines format in which one record is one text from one perspective (writer, experiencer,
reader, or not specified; annotators' reconstructions of the writer's emotion count as writer). The original labels are kept; emotion categories are
additionally mapped to a shared set (anger, disgust, fear, joy,
sadness, surprise, shame/guilt, trust, no emotion); the website shows
this mapping as a tree of all \prelim{130} original labels, and labels
outside the shared set (e.g., love, pride, relief) are kept and
predicted by models trained on all labels of one corpus.
Valence--arousal, appraisals, emotion components, and role spans are
kept where a corpus has them. Each corpus has a fixed split: the
official split where we use one, a split grouped by author, story or
scenario where texts depend on each other, and otherwise 80/10/10 by a
hash of the normalised text, so that identical texts share a split
across corpora; the page of each corpus states
which. %
Our own corpora can be downloaded in this format where their licenses
allow it; the corpora of other groups are linked to their original
source.\footnote{Depending on the feedback of resource authors that we
  include, we adapt the availability.}

\begin{figure}[t]
  \centering
  \begin{tikzpicture}[
      box/.style={draw, rounded corners=2pt, font=\scriptsize\sffamily, align=center, execute at begin node={\hyphenpenalty=10000\relax}, inner sep=2pt, minimum height=0.55cm},
      head/.style={box, fill=black!5, text width=1.15cm, minimum height=0.75cm},
      arr/.style={-{Stealth[length=1.5mm]}, thin}]
    \node[box, text width=6.9cm] (in) at (0,0) {perspective token (perspective model) or fixed \texttt{[not-specified]} (no perspective model) + text};
    \node[box, fill=black!12, text width=6.9cm] (enc) at (0,-0.85) {shared encoder (XLM-R base), first-token representation};
    \node[head] (cat)  at (-2.86,-1.95) {categories};
    \node[head] (va)   at (-1.43,-1.95) {valence, arousal};
    \node[head] (gate) at (0,-1.95)     {event gate};
    \node[head] (app)  at (1.43,-1.95)  {appraisals};
    \node[head] (cpm)  at (2.86,-1.95)  {emotion\\components};
    \draw[arr] (in) -- (enc);
    \foreach \h in {cat,va,gate,app,cpm} \draw[arr] (enc.south -| \h) -- (\h.north);
    \node[box, text width=2.3cm] (out) at (1.43,-3.05) {appraisals output only for events};
    \draw[arr] (app) -- (out);
    \draw[arr, dashed] (gate.south) |- (out.west);
    \node[box, dashed, text width=6.9cm] (role) at (0,-3.95) {separate token tagger: experiencer, stimulus, target and cue spans};
  \end{tikzpicture}
  \caption{The general model of the \hub{}. One encoder is trained jointly on all corpora with human annotations. A token precedes each text: its perspective in the perspective model, the same
    fixed token in the no perspective model. Each corpus contributes loss only to the heads and
    labels it annotates. The event gate decides whether appraisals are output. Roles come from a separate span tagger.}
  \label{fig:model}
\end{figure}
 
\paragraph{Models.}
For each corpus with emotion categories, we fine-tune XLM-RoBERTa base
and DeBERTa-v3 base (English only; the English part of \res{BRIGHTER},
\res{XED} and \res{PO-EMO}) with three seeds, and train a bag-of-words
one-vs-all logistic regression baseline; all use a class-weighted loss
and per-label thresholds tuned on the validation data (details on the
website). For each corpus with valence--arousal, appraisal, or
component annotations, we train additional models. Models trained on
combinations of corpora cover one kind of text (domain models) or all
corpora with human annotations (general model; without automatically
labelled corpora such as \res{TEC}). The general model
(Figure~\ref{fig:model}) shares one encoder with heads for categories,
valence--arousal, appraisals, and emotion components, and an event
gate that decides whether a text describes an event, so that
appraisals are only output where they are meaningful; gate and
component head also learn from our own annotations. We train it in two
variants: In the \emph{perspective model}, a token (\texttt{[writer]},
\texttt{[experiencer]}, \texttt{[reader]} or \texttt{[not-specified]})
precedes each text and states whose emotion is meant. In the \emph{no
  perspective model}, every text receives the same token
(\texttt{[not-specified]}) in training and use. Since the perspective
of new texts is often unknown, we use the no-perspective model for the
case studies, and ``general model'' refers to it unless stated
otherwise. Role labeling models tag experiencers, stimuli, targets and
cues as spans; one tagger per role corpus serves the cross-corpus
evaluation and one joint English tagger serves applications. Zero-shot
prompted LLMs (five 7--12B models, Llama~3.3~70B, Qwen3~32B and
gpt-oss-120b) further contribute to the cross-corpus evaluation.

\paragraph{Cross-corpus evaluation.}
Every model is tested on the test split of every corpus; texts that occur in both the training and the test corpus
are removed, for the two variants of the general model also all texts used to train or tune their event gate. Users pick the set of emotions on which the models are compared (by default anger, fear, joy and sadness);
each cell shows the macro-F$_1$ on these emotions, and corpus pairs that do not both have them are not shown in the
heatmap. Alternatively, a cell shows the transfer ratio, i.e., the macro-F$_1$ of a model on a test corpus divided by
that of the test corpus' own model on the same emotions. The findings in Section~\ref{sec:findings} use all emotions
that the two corpora of a cell share.

\paragraph{Website.}
The website has five sections. The landing page summarises the other four. \emph{Data sets} lists all
corpora (filterable by origin, kind of text, language and annotation) with label shares; each corpus has a page with a description, an annotated example, label statistics, characteristic words, the split, the license, similar corpora and its models; a similarity matrix and the tree of all original labels complete the section. \emph{Models} offers the general, domain and corpus-specific models with code for
their use with the \texttt{transformers} library, and the cross-corpus results as heatmaps (for all representations, with a choice of models, rows, columns and labels). \emph{Application} contains a
demo and a batch processing page, which run the role tagger and, at the user's choice, the no-perspective model,
the perspective model (with a choice of perspective, by default the writer) or a categories-only model trained on 30 corpora,
including automatically labelled ones, in the browser without sending texts
to a server, and the case studies (Section~\ref{sec:casestudies}) with aggregates, subsets and a word search for each
data set. \emph{Additional information} holds our course on emotion analysis, technical details, ethical
considerations and a glossary.

\section{Findings}
\label{sec:findings}
Table~\ref{tab:findings} summarises the results of this section. Unless stated otherwise, all numbers for emotion
categories refer to the shared emotion labels only.

\subsection{What Works}
\label{sec:works}
\textit{In-corpus classification.} Fine-tuned XLM-R reaches a mean in-corpus macro-F$_1$ of \prelim{.52} over the
\prelim{39} corpus views with emotion categories (DeBERTa \prelim{.54} on the English ones), but the corpora differ widely.
Models score highest where the labels follow from the text by construction or by distant supervision:
on \res{Emotion-Stimulus} (constructed sentences, \prelim{.97}), \res{CARER} (tweets labelled by their hashtags,
\prelim{.91}) and \res{Tales} (five labels on sentences with high annotator agreement, \prelim{.81}). Self-reports of
events, as in \res{ISEAR} and \res{EmpatheticDialogues}, follow with \mbox{\prelim{.75--.76}}, \res{crowd-enVent} with masked emotion words with \prelim{.59}. Conversations
(\res{DailyDialog}, \res{MELD}: \mbox{\prelim{.39--.40}}), news headlines, arguments and reader annotations are hard, down to
\prelim{.17} for the readers of \res{GoodNews\-Everyone}.

\textit{Transfer.} A model trained on one corpus reaches a median of \prelim{65\%} of the target corpus' own model
when both contain the same kind of text (XLM-R; DeBERTa \prelim{67\%}, bag-of-words \prelim{61\%}). In \prelim{56}
cells, a model trained on another corpus is even better than the test corpus' own model, mostly on small test
corpora such as \res{EmoTales} and \res{EmoProgress}. Valence and arousal transfer best: dimensional models keep a median of \prelim{86\%}
(XLM-R) and \prelim{90\%} (DeBERTa) of the target corpus' own model.

\textit{One model for all corpora.} The no-perspective model, our general model, comes close to the single-corpus
models on its training corpora: on the \prelim{27} views with human annotations, it reaches a mean macro-F$_1$ of \prelim{.514} (the corpora's
own XLM-R models: \prelim{.523}), with valence--arousal $r$ \prelim{.69}, core appraisals $r$ \prelim{.53} and components
F$_1$ \prelim{.56}. It beats the single-corpus models on small appraisal corpora but stays below them on the two component
corpora. Not distinguishing perspectives costs little: with the correct tokens, the perspective model reaches \prelim{.526}
on these views.

\textit{Zero-shot LLMs on the basic emotions.} On anger, fear, joy and sadness, prompted Llama~3.3~70B reaches a mean
macro-F$_1$ of \prelim{.65} over the \prelim{39} test views, slightly above each corpus' own fine-tuned XLM-R model
(\prelim{.63}), and it is better on \prelim{25} of them; gpt-oss-120b (\prelim{.62}) and Qwen3~32B (\prelim{.59}) are close,
and the five 7--12B LLMs (Gemma~3, Qwen3, Granite~3.3, Llama~3.1, Mistral) reach \mbox{\prelim{.57--.62}}.
This comes at a much higher cost: on the case-study texts, gpt-oss-120b labels about \prelim{50} texts per second and
Llama~3.3~70B about \prelim{6} (four A100 GPUs), the no-perspective model (\prelim{278M} parameters) about
\prelim{1,200--1,400} on one consumer GPU, with all representations at once.

\subsection{Where Emotion Analysis Needs Work}
\label{sec:needswork}
\textit{Generalisation across domains and languages.} When the training and test corpus contain different kinds of
text, the median transfer ratio drops to \prelim{53\%} (XLM-R; DeBERTa \prelim{57\%}). Models with a higher in-corpus
score transfer only slightly better: across kinds of English text, bag-of-words keeps
\prelim{51\%}, XLM-R \prelim{55\%} and DeBERTa \prelim{57\%}. Between English and German,
XLM-R keeps \prelim{48\%} from German to English and \prelim{47\%} in the other direction.

\textit{Appraisals and emotion components.} Appraisal models keep only \prelim{64\%} (XLM-R) and \prelim{61\%} (DeBERTa) of the target corpus' own
model. Only two corpora have emotion components; their
models keep a median of \prelim{78\%} (XLM-R) and \prelim{73\%} (DeBERTa) and transfer better from tweets to literature (\prelim{86--89\%}) than in the other
direction (\prelim{60--68\%}).

\begin{table}[t]
  \centering\small
  \begin{tabular}{@{}lcc@{}}
    \toprule
    & XLM-R & DeBERTa \\
    \midrule
    \multicolumn{3}{@{}l}{\textbf{In-corpus macro-F$_1$}} \\
    Mean over corpora & \prelim{.52} & \prelim{.54} \\
    Lowest corpus & \prelim{.17} & \prelim{.17} \\
    Highest corpus & \prelim{.97} & \prelim{.97} \\
    Mean by perspective: & & \\
    \quad writer & \prelim{.55} & \prelim{.58} \\
    \quad experiencer & \prelim{.54} & \prelim{.56} \\
    \quad not specified & \prelim{.53} & \prelim{.56} \\
    \quad reader & \prelim{.29} & \prelim{.29} \\
    \midrule
    \multicolumn{3}{@{}l}{\textbf{Transfer ratio (\%)}} \\
    Same kind of text & \prelim{65} & \prelim{67} \\
    Other kind of text & \prelim{53} & \prelim{57} \\
    German $\rightarrow$ English & \prelim{48} & -- \\
    English $\rightarrow$ German & \prelim{47} & -- \\
    Valence--arousal & \prelim{86} & \prelim{90} \\
    Appraisals & \prelim{64} & \prelim{61} \\
    Emotion components$^*$ & \prelim{78} & \prelim{73} \\
    \midrule
    \multicolumn{3}{@{}l}{\textbf{General model (\% of single-corpus models)}} \\
    On its training corpora & \prelim{97--99} & -- \\
    On an unseen corpus$^\dagger$ & \prelim{69} & -- \\
    \bottomrule
  \end{tabular}
  \caption{Overview of the findings; emotion categories on the shared labels. In-corpus: macro-F$_1$ of each corpus' own
    model. Transfer ratio: score on another corpus in \% of that corpus' own model (macro-F$_1$ for categories and
    components, Pearson's $r$ for valence--arousal and appraisals). DeBERTa is English only. General model: the no-perspective model (XLM-R). $^*$Two corpus pairs. $^\dagger$On validation data.}
  \label{tab:findings}
\end{table}

\textit{Perspective.} In-corpus macro-F$_1$ is similar for the writer (\prelim{.55}), the experiencer
(\prelim{.54}) and unspecified perspectives (\prelim{.53}); reader emotions are the hardest (\prelim{.29}). A model for one perspective does not simply serve another: on \res{GoodNews\-Everyone}, where readers react to
news, the experiencer model keeps \prelim{49\%} of the reader model and the reader model \prelim{37\%} of the
experiencer model.

\textit{LLMs beyond the basic emotions.} On all labels a corpus shares with the hub's set, the three large LLMs reach a
median of \prelim{89--92\%} of the corpus' own XLM-R model (better on \prelim{11--13} of \prelim{39} views), the 7--12B LLMs \prelim{80--86\%}. The gap is largest for \emph{no emotion} (F$_1$
\prelim{.17--.26} lower), \emph{trust} (\prelim{.05--.12}) and \emph{surprise} (\prelim{.07--.08}). What counts as no emotion
or surprise differs between corpora, and a prompt hardly conveys this. The 32--120B models gain only \prelim{2--3} points
over the best 8--12B model (Qwen3~8B, \prelim{.43}).

\textit{Unseen corpora.} Without one of its training corpora, the no-perspective model drops from \prelim{89\%} to \prelim{69\%}
of that corpus' own model (validation data), above a typical single-corpus model (\prelim{54\%}, test).

\textit{When to predict appraisals.} Annotators agree only moderately on whether a text describes an event
(\mbox{$\kappa$ \prelim{.43}}); the no-perspective model's gate reaches a validation AUC of \prelim{.89}.

\section{Case Studies}
\label{sec:casestudies}
Six case studies show how far the hub's models answer questions from the social sciences, literary studies and health
research. Each uses a full, public English data set and at least two tasks; three are described below, three in
Appendix~\ref{app:cs}, and the website has a page for each.

\paragraph{Setup.}
All studies apply the released models without further training: the no-perspective general model, as the perspective
of these texts is unknown (appraisals only for texts that the event gate marks as events), the domain model of the kind
of text as a robustness check, and the role tagger. We report
aggregates only, with 95\% bootstrap intervals. On \prelim{300} manually annotated texts, the general model reaches a
macro-F$_1$ of \prelim{.35} (per source \prelim{.19--.35}).

\paragraph{CS1: What do people bring to language models?}
We want to know which emotions users express in their requests to chat assistants, and whether users who ask for
personal advice appraise the events they describe differently from users with other requests (we expected lower
control and higher self-responsibility). The data are the first user turns of the English conversations in WildChat-1M
(\prelim{477k}; \citealp{zhao-etal-2024-wildchat}) and LMSYS-Chat-1M (\prelim{777k}; \citealp{zheng-etal-2024-lmsys}). A zero-shot LLM (Qwen3~32B) assigns each turn a request
type; on \prelim{200} turns annotated by the author it agrees in \prelim{87.5\%} of the cases ($\kappa$ \prelim{.83}), but finds
only \prelim{22\%} of the advice requests. Coding and information
requests express no emotion in \prelim{97--99\%} of the cases; advice requests express anger (\prelim{12\%}),
sadness (\prelim{10\%}) and fear (\prelim{8\%}); creative-writing requests carry the emotions of the requested text (joy
\prelim{18\%}, fear \prelim{9\%}). Compared with coding and information requests, advice requests are appraised as less
certain in WildChat and LMSYS (\prelim{$-.28$}, \prelim{$-.08$}; 0--1 scale), in all four variants of the general model that we compared.
Our expectations on control and self-responsibility find no robust support: in WildChat, the differences are small (self-responsibility \prelim{$-.06$}, own control \prelim{$-.02$}) and change sign across the variants.

\paragraph{CS2: Showing or telling emotions in fiction?}
We want to know whether fiction conveys emotions more by showing them (bodily symptoms, expressions) or by telling them
(named feelings, appraisals), and how this changed; we expected nineteenth-century fiction to show more and later fiction
to tell more. The data are \prelim{1.76M} paragraphs of \prelim{1,690} English novels from Project Gutenberg by authors born
1760--1899; lacking publication dates, we order the books by their author's birth year. We compute, per birth decade, the share of emotional paragraphs with each predicted component and its trend (books resampled). The general model finds an emotion in \prelim{82\%} of the
paragraphs (literature domain model: \prelim{65\%}). Across birth decades, the share with subjective feeling falls by
\prelim{0.6} points per decade (CI \prelim{[$-0.7$, $-0.5$]}), as does motivation (\prelim{$-0.6$}), while motor expression (\prelim{+0.6}) and bodily symptoms (\prelim{+0.2}) rise slightly; cognitive appraisal falls (\prelim{$-0.3$}), but changes sign across model variants: largely the opposite of our expectation and in line with the ``show, don't tell'' convention of
later novels.

\paragraph{CS3: How much do the hub's models agree in the wild?}
We want to know how consistent the hub's models are on texts unlike any training corpus, and whether the cross-corpus
results predict which models agree. The data are \prelim{2,000} random texts from each of the seven case-study
sources. The general model, the domain model of each source, the \prelim{39} single-corpus XLM-R models, an external
classifier \citep{hartmann-2022-emotion} and the six prompted LLMs label all of them; we compute Cohen's $\kappa$ per label
on the labels two models share and average over labels and sources. Models trained on different corpora barely agree
(median \prelim{.07}); the general model agrees more with them (\prelim{.15}), with the domain models (\prelim{.25}), the
external classifier (\prelim{.30}) and the LLMs (\prelim{.28}); the LLMs agree most with each other (\prelim{.49}). The models disagree already on whether a text expresses any emotion: the general model assigns no
emotion to \prelim{69\%} of the ABC headlines, the news domain model to \prelim{1\%}, Llama~3.3~70B to \prelim{23\%}. The
cross-corpus evaluation predicts this agreement: a model's mean transfer ratio correlates with its mean agreement with the other single-corpus models on the new texts
(Spearman $\rho$ \prelim{.89}, $n=39$, $p<.001$), and for pairs of models the transfer ratio correlates with agreement
($\rho$ \prelim{.45}, Mantel test $p<.001$). The heatmaps thus indicate which models are more consistent on new data,
although agreement is not correctness.

\section{Conclusion}
\label{sec:conclusion}
The \hub{} brings together \prelim{48} emotion corpora, models for all representations, and their
cross-corpus evaluation.
\textbf{RQ1:} Models generalise within a kind of text (median \prelim{65\%} of the target's own model) but less
across kinds of text (\prelim{53\%}) and languages (\prelim{47--48\%}); valence and arousal transfer well, appraisals
less so, and experiencer and reader models do not replace each other.
\textbf{RQ2:} A single general model, with or without perspectives, comes close to the single-corpus models on their own corpora
(below them on emotion components), but on texts no corpus covers its quality drops (macro-F$_1$ \prelim{.35} on our
case-study annotations). Prompted LLMs match fine-tuned models on the basic emotions but not on others.
\textbf{RQ3:} At scale, the models' predictions are in line with appraisal theories (Appendix~\ref{app:cs};
possibly learned from corpora with both annotations), largely contradict two of our hypotheses (more ``telling'' in later fiction, more anger than fear in negative
reviews) and give mixed evidence for a third (news: less fear and anger in ABC, more in HuffPost). Users should pick a model for their kind of text, check its transfer ratios and validate it
on a sample.

\section*{Limitations}
The hub covers English and German only, and text only: several corpora come with images, audio or video, which we do not
use. Mapping the labels of all corpora to one shared set is an editorial choice:
merged labels (shame and guilt, synonyms such as \emph{annoyance}) and unmapped labels (kept only in the full-label
models) change what a cross-corpus cell measures. Both the mapping and the selection of the shared labels can be debated,
depending on the underlying emotion theory. The transfer ratio depends on the quality of the test corpus' own
model; on small corpora this model is weak, so ratios above 100\% occur. Other groups' corpora are not redistributed,
so results on them can only be reproduced after downloading them from their original sources. The no-perspective model does not distinguish whose emotion is meant, so its outputs cannot be read as the emotion of a particular person. The models inherit the
biases of their training corpora, e.g., in the demographics of crowd workers and authors. The event gate is tuned for
recall; on reviews and Reddit posts it marks almost every text (\prelim{98--100\%}) as an event and thus hardly filters.
It was trained together with the shared encoder on \prelim{802} texts, \prelim{327} of which come from the test splits of the
hub's corpora; we exclude all texts used to train or tune the gate from the test results of both general-model variants, but
this cannot rule out an indirect influence on these results.
The case studies apply the models outside their training distribution; they are descriptive, use simple operationalisations
(a person lexicon for stimuli, the author's birth year instead of the publication year, an LLM for request types that
misses most advice requests), and their results describe what the models predict.

\section*{Ethics Statement}
Emotion recognition can be misused, e.g., for surveillance or manipulation \citep{mohammad-2022-ethics-sheet}; the hub
is meant for research on texts, not for inferring the emotion state of individuals. The demo and batch processing run
in the browser and do not send or store user texts. We redistribute only corpora whose licenses allow it (most of our own under
CC~BY~4.0, where all authors agreed) and link to the original sources of all others; models trained on corpora with
non-commercial licenses carry these restrictions. The released data contain no worker IDs or demographic information.
The case-study data are public; we analyse them only in aggregate, show no example texts from user-generated data, and
make no claims about individuals, in particular not for the health and mental-health data.

\section*{Acknowledgements}
We thank all co-authors of the resources included in the hub and all colleagues working on emotion analysis, in
particular Jeremy Barnes, Aidan Combs, Rossana Damiano, Bao Minh Doan Dang, Lynn Greschner, Amelie Heindl, Anna Khlyzova, Laura Oberl\"ander, Sebastian Pad\'o, Viviana Patti, Hendrik Schuff, Carina Silberer, Marco Antonio Stranisci, Enrica Troiano and Sabine Weber for their contributions. %
This work
has been supported by the projects EMCONA (The Interplay of Emotions and Convincingness in Arguments, KL 2869/12-1,
project number 516512112), INPROMPT (Interactive Prompt Optimization with the Human in the Loop for Natural Language
Understanding Model Development and Intervention, KL 2869/13-1, project number 521755488) and ITEM (User's Choice of
Images and Text to Express Emotions in Twitter and Reddit, KL 2869/11-1, project number 513384754), all funded by the
German Research Foundation (DFG).
We used Claude (Anthropic) to write code and analysis scripts and to draft text; all text was checked and revised by
the author. In an experiment, Claude also annotated \prelim{1,000} texts for emotion components; these labels are not
used in the released models, which are trained only on human annotations. Request types in CS1 were assigned by
Qwen3~32B.

\clearpage
\appendix

\section{Further Case Studies}
\label{app:cs}
\paragraph{CS4: Who makes the news frightening?}
We want to know whether news headlines became more frightening and angrier and whether their stimuli increasingly name
persons (personalisation of news); we expected both. The data are the Australian ABC headlines of 2003--2021
(\prelim{1.24M}; \citealp{kulkarni-2018-million}) and the HuffPost headlines of 2012--2022 (\prelim{210k};
\citealp{misra-2022-news}). The general model predicts fear and anger for each
headline (the news domain model as a check), and we compute yearly shares and their trend; for HuffPost, we also weight
every year by the overall mix of sections. Stimulus spans from the role tagger count as naming a person if they contain a
person noun or the name of a head of government (ABC headlines are lower-cased, so NER does not work). In ABC, the share
of fear and anger fell by \prelim{3.1} points per decade (\prelim{27.9\%} in 2003, \prelim{24.6\%} in 2021; domain model
\prelim{$-3.2$}). In HuffPost, it rose from \prelim{12\%} to \prelim{38\%}, and within the section mix from \prelim{23\%} to
\prelim{30\%} (\prelim{+10.0} points per decade, CI \prelim{[8.4, 11.5]}). Stimuli name persons slightly more often in ABC
(\prelim{+2.5} points per decade) and clearly more often in HuffPost (\prelim{17\%} in 2012, \prelim{27--33\%} in 2016--20).

\paragraph{CS5: Anger and fear in patient reviews.}
We want to know which emotions patients express in negative reviews of drugs; we expected more anger, attributed to
others, than fear. The data are \prelim{215k} reviews from Drugs.com with ratings \citep{grasser-etal-2018-aspect}. The general
model predicts the categories of each review and the appraisals of reviews that describe an event; the role tagger
finds stimulus spans, which we assign to drugs, side effects or care providers by lexicons. In reviews rated 1--4, the
general model finds sadness in \prelim{48\%}, fear in \prelim{46\%} and anger in \prelim{14\%} (domain model: fear
\prelim{27\%}, anger \prelim{3\%}). Anger reviews are appraised as caused (\prelim{+.35}) and controlled (\prelim{+.31}) by
others and as more certain (\prelim{+.45}) than fear reviews, in line with appraisal theories, and their stimuli name care
providers almost twice as often (\prelim{5.8\%} vs.\ \prelim{3.0\%}), while side effects are about as frequent (\prelim{15\%} vs.\
\prelim{17\%}).

\paragraph{CS6: Appraisals in anxiety and depression communities.}
We want to know whether people in mental-health communities appraise the events they describe as appraisal and
attribution theories predict: with low certainty and control in anxiety communities, and with low pleasantness and high
self-responsibility in depression communities. The data are \prelim{441k} posts of the Reddit Mental Health Dataset
\citep{low-etal-2020-natural} from r/Anxiety, r/HealthAnxiety, r/depression and r/lonely, and from r/fitness,
r/personalfinance and r/relationships as control communities. The general model predicts categories and the appraisals
of posts that describe an event, which we compare with the control communities; the role tagger separates posts whose
experiencer is the writer. Fear is predicted for \prelim{74\%} of the posts in r/Anxiety and \prelim{85\%} in r/HealthAnxiety,
sadness for \prelim{75\%} in r/depression and \prelim{72\%} in r/lonely (control communities: fear \prelim{11--18\%}). Anxiety
posts are appraised as less certain (\prelim{$-.18$}) and less controllable (\prelim{$-.18$}), as expected; depression posts
as less pleasant (\prelim{$-.12$}) and less controllable (\prelim{$-.17$}), but hardly as more self-caused (\prelim{+.00}). The
same holds for posts whose experiencer is the writer. We make no claims about individuals.

\section{Corpora}
\label{app:corpora}
Table~\ref{tab:corpora} lists all corpora in the \hub{}.

\begin{table*}[p]
  \centering\fontsize{6}{7.2}\selectfont
  \renewcommand{\arraystretch}{0.93}\setlength{\tabcolsep}{3pt}
  \begin{tabularx}{\textwidth}{p{4.2cm}p{5cm}X}
  \toprule
  Corpus (reference) & Description & Example \\
  \cmidrule(r){1-1}\cmidrule(rl){2-2}\cmidrule(l){3-3}
  \raggedright\href{http://web.eecs.umich.edu/\~{}mihalcea/affectivetext/}{Affective Text} \citep{strapparava-mihalcea-2007-semeval} & News headlines; intensity of six emotions and valence. & ``Town packs gym to mourn 10 killed in fire'' \\
  \arrayrulecolor{black!25}\specialrule{0.2pt}{2pt}{2pt}\arrayrulecolor{black}
  \raggedright\href{https://saifmohammad.com/WebPages/affectintweets.htm}{AIT} \citep{mohammad-etal-2018-semeval} & Tweets with 11 emotions, valence and intensity. & ``Might just leave and aggravate bae'' \\
  \arrayrulecolor{black!25}\specialrule{0.2pt}{2pt}{2pt}\arrayrulecolor{black}
  \raggedright\href{https://www.romanklinger.de/data-sets/}{Appraisal strategies} \citep{hofmann-etal-2021-emotion} & enISEAR/deISEAR events, appraisals by several strategies. & ``I felt \dots{} when I watched videos of how some people treat animals.'' \\
  \arrayrulecolor{black!25}\specialrule{0.2pt}{2pt}{2pt}\arrayrulecolor{black}
  \raggedright\href{https://www.romanklinger.de/data-sets/}{Appraisal-enISEAR} \citep{hofmann-etal-2020-appraisal} & enISEAR events with seven appraisal dimensions. & ``I felt \dots{} watching my sapphire ring sparkle in the sunshine.'' \\
  \arrayrulecolor{black!25}\specialrule{0.2pt}{2pt}{2pt}\arrayrulecolor{black}
  \raggedright\href{https://huggingface.co/datasets/marco-stranisci/APPreddit}{APPReddit} \citep{stranisci-etal-2022-appreddit} & Events in Reddit posts with five appraisal dimensions. & ``I just need to know that it's going to be ok one day.'' \\
  \arrayrulecolor{black!25}\specialrule{0.2pt}{2pt}{2pt}\arrayrulecolor{black}
  \raggedright\href{https://huggingface.co/brighter-dataset}{BRIGHTER} \citep{muhammad-etal-2025-brighter} & Texts of several genres, emotions and intensities (en, de). & ``our hands outstretched for the most amazing high five of my life.'' \\
  \arrayrulecolor{black!25}\specialrule{0.2pt}{2pt}{2pt}\arrayrulecolor{black}
  \raggedright\href{https://huggingface.co/datasets/dair-ai/emotion}{CARER} \citep{saravia-etal-2018-carer} & Tweets labelled distantly by emotion hashtags. & ``i am on this track i feel good things coming'' \\
  \arrayrulecolor{black!25}\specialrule{0.2pt}{2pt}{2pt}\arrayrulecolor{black}
  \raggedright\href{https://www.uni-bamberg.de/en/nlproc/projects/emcona/}{ContArgA} \citep{greschner-etal-2026-trust} & Arguments with readers' emotions and appraisals. & ``naturopathy may have no scientific evidence.'' \\
  \arrayrulecolor{black!25}\specialrule{0.2pt}{2pt}{2pt}\arrayrulecolor{black}
  \raggedright\href{https://www.uni-bamberg.de/en/nlproc/projects/emcona/}{Context-ContArgA} \citep{greschner-etal-2026-trust} & ContArgA arguments in their situation, one record per reader. & ``An old woman brings up an argument to another person. [\dots]'' \\
  \arrayrulecolor{black!25}\specialrule{0.2pt}{2pt}{2pt}\arrayrulecolor{black}
  \raggedright\href{https://huggingface.co/datasets/coping-emotions/roleplay-corpus}{COPING} \citep{troiano-etal-2024-dealing} & Replies in role-played confrontations, own emotions. & ``[\dots] I don't see why anyone would want to use drugs though.'' \\
  \arrayrulecolor{black!25}\specialrule{0.2pt}{2pt}{2pt}\arrayrulecolor{black}
  \raggedright\href{https://github.com/honglizhan/CovidET-Appraisals-Public}{CovidET-Appraisals} \citep{zhan-etal-2023-evaluating} & Reddit posts on COVID-19 with 24 appraisal dimensions. & ``Lately, I keep trying to distract myself [\dots]'' \\
  \arrayrulecolor{black!25}\specialrule{0.2pt}{2pt}{2pt}\arrayrulecolor{black}
  \raggedright\href{https://www.romanklinger.de/data-sets/}{CPM-REMAN} \citep{casel-etal-2021-emotion} & REMAN passages with five emotion components. & ``No sound reached my ear. I felt an unbearable uneasiness. [\dots]'' \\
  \arrayrulecolor{black!25}\specialrule{0.2pt}{2pt}{2pt}\arrayrulecolor{black}
  \raggedright\href{https://www.romanklinger.de/data-sets/}{CPM-TEC} \citep{casel-etal-2021-emotion} & TEC tweets with five emotion components. & ``Turns out the machine can't figure out my problem and now I'm on hold'' \\
  \arrayrulecolor{black!25}\specialrule{0.2pt}{2pt}{2pt}\arrayrulecolor{black}
  \raggedright\href{https://www.romanklinger.de/data-sets/}{crowd-enVent} \citep{troiano-etal-2023-dimensional} & Events with writers' emotions and appraisals (masked text). & ``My child was rushed to hospital after a freak accident'' \\
  \arrayrulecolor{black!25}\specialrule{0.2pt}{2pt}{2pt}\arrayrulecolor{black}
  \raggedright\href{https://data.world/crowdflower/sentiment-analysis-in-text}{CrowdFlower} \citep{crowdflower-2016-emotion} & Tweets with 13 crowd-sourced emotion labels. & ``may the 4th be with you! HAPPY STAR WARS DAY!'' \\
  \arrayrulecolor{black!25}\specialrule{0.2pt}{2pt}{2pt}\arrayrulecolor{black}
  \raggedright\href{http://yanran.li/dailydialog.html}{DailyDialog} \citep{li-etal-2017-dailydialog} & Utterances of everyday dialogues, six emotions or none. & ``Oh yes and try lots of new food too.'' \\
  \arrayrulecolor{black!25}\specialrule{0.2pt}{2pt}{2pt}\arrayrulecolor{black}
  \raggedright\href{https://www.romanklinger.de/data-sets/}{deISEAR} \citep{troiano-etal-2019-crowdsourcing} & German events for prompted emotions. & ``Ich fühlte \dots{}, als mich ein Arbeitskollege beleidigt hat.'' \\
  \arrayrulecolor{black!25}\specialrule{0.2pt}{2pt}{2pt}\arrayrulecolor{black}
  \raggedright\href{https://www.uni-bamberg.de/en/nlproc/projects/item/}{Donate or Create} \citep{bagdon-etal-2025-donate} & Social media posts, emotions, appraisals. & ``Going home tomorrow, last night of holiday - so sad to leave'' \\
  \arrayrulecolor{black!25}\specialrule{0.2pt}{2pt}{2pt}\arrayrulecolor{black}
  \raggedright\href{https://saifmohammad.com/WebPages/ResearchInterests.html}{Electoral Tweets} \citep{mohammad-etal-2015-electoral} & Tweets on the 2012 US presidential election. & ``Mitt Romney Vs Barack Obama [\dots] BRING IT ON'' \\
  \arrayrulecolor{black!25}\specialrule{0.2pt}{2pt}{2pt}\arrayrulecolor{black}
  \raggedright\href{https://github.com/JULIELab/EmoBank}{EmoBank} \citep{buechel-hahn-2017-emobank} & Sentences of several genres, writer and reader VAD. & ``Hong Kong's state-of-the-art interactive museums [\dots]'' \\
  \arrayrulecolor{black!25}\specialrule{0.2pt}{2pt}{2pt}\arrayrulecolor{black}
  \raggedright\href{https://github.com/fmplaza/EmoEvent}{EmoEvent} \citep{plaza-del-arco-etal-2020-emoevent} & Tweets about events of 2019 with Ekman's emotions. & ``A glorious week for all the starks. \#Endgame \#GameOfThrones'' \\
  \arrayrulecolor{black!25}\specialrule{0.2pt}{2pt}{2pt}\arrayrulecolor{black}
  \raggedright\href{https://saifmohammad.com/WebPages/EmotionIntensity-SharedTask.html}{EmoInt} \citep{mohammad-bravo-marquez-2017-wassa} & Tweets with the intensity of anger, fear, joy or sadness. & ``The new lock sound is grim'' \\
  \arrayrulecolor{black!25}\specialrule{0.2pt}{2pt}{2pt}\arrayrulecolor{black}
  \raggedright\href{https://lt3.ugent.be/resources/emoprogress/}{EmoProgress} \citep{wemmer-etal-2024-emoprogress} & Dreams and service dialogues, emotions, appraisals. & ``There was a strange game thing with an evil snake creature. [\dots]'' \\
  \arrayrulecolor{black!25}\specialrule{0.2pt}{2pt}{2pt}\arrayrulecolor{black}
  \raggedright\href{https://github.com/emorynlp/emotion-detection}{EmoryNLP} \citep{zahiri-choi-2018-emotion} & Utterances from the TV series Friends, seven emotions. & ``Oh, can I borrow this? My milk's gone bad.'' \\
  \arrayrulecolor{black!25}\specialrule{0.2pt}{2pt}{2pt}\arrayrulecolor{black}
  \raggedright\href{https://doi.org/10.1007/s10579-011-9140-5}{EmoTales} \citep{francisco-etal-2012-emotales} & Sentences of folk tales with emotions and VAD. & ``It was really only a little linnet singing outside his window, [\dots]'' \\
  \arrayrulecolor{black!25}\specialrule{0.2pt}{2pt}{2pt}\arrayrulecolor{black}
  \raggedright\href{https://www.romanklinger.de/data-sets/}{Emotion channels} \citep{kim-klinger-2019-analysis} & Fan fiction, characters' emotions and their channels. & ``Embry Call had been on cloud nine for four days now. [\dots]'' \\
  \arrayrulecolor{black!25}\specialrule{0.2pt}{2pt}{2pt}\arrayrulecolor{black}
  \raggedright\href{http://www.site.uottawa.ca/\~{}diana/resources/emotion_stimulus_data/}{Emotion-Stimulus} \citep{ghazi-etal-2015-detecting} & Constructed sentences with emotion and stimulus spans. & ``The guard stiffened in anger.'' \\
  \arrayrulecolor{black!25}\specialrule{0.2pt}{2pt}{2pt}\arrayrulecolor{black}
  \raggedright\href{https://github.com/facebookresearch/EmpatheticDialogues}{EmpatheticDialogues} \citep{rashkin-etal-2019-towards} & Situations described by speakers for one of 32 emotions. & ``I miss the days of youth where we were carefree'' \\
  \arrayrulecolor{black!25}\specialrule{0.2pt}{2pt}{2pt}\arrayrulecolor{black}
  \raggedright\href{https://www.romanklinger.de/data-sets/}{enISEAR} \citep{troiano-etal-2019-crowdsourcing} & English events for prompted emotions. & ``I felt \dots{} when I saw somebody kicking their dog.'' \\
  \arrayrulecolor{black!25}\specialrule{0.2pt}{2pt}{2pt}\arrayrulecolor{black}
  \raggedright\href{https://github.com/wwbp/additional_data_sets/tree/master/valence_arousal}{Facebook VA} \citep{preotiuc-pietro-etal-2016-modelling} & Facebook posts with valence and arousal. & ``I am on cloud nine right now.'' \\
  \arrayrulecolor{black!25}\specialrule{0.2pt}{2pt}{2pt}\arrayrulecolor{black}
  \raggedright\href{https://www.romanklinger.de/data-sets/}{Fan-fiction relations} \citep{kim-klinger-2019-frowning} & Fan fiction, emotional relations between characters. & ``[\dots] Cortana rolled her eyes at his excuse. [\dots]'' \\
  \arrayrulecolor{black!25}\specialrule{0.2pt}{2pt}{2pt}\arrayrulecolor{black}
  \raggedright\href{https://doi.org/10.7910/DVN/C9SAIX}{German political texts} \citep{widmann-wich-2023-creating} & German parliamentary speeches and party Facebook posts. & ``Warum herrschen beim DFB Cliquen und Klüngel?'' \\
  \arrayrulecolor{black!25}\specialrule{0.2pt}{2pt}{2pt}\arrayrulecolor{black}
  \raggedright\href{https://www.romanklinger.de/data-sets/}{GerSti} \citep{doan-dang-etal-2021-emotion} & German news headlines with emotions and stimulus spans. & ``30 Jahre Deutsche Einheit: Nicht immer nur jammern!'' \\
  \arrayrulecolor{black!25}\specialrule{0.2pt}{2pt}{2pt}\arrayrulecolor{black}
  \raggedright\href{https://github.com/google-research/google-research/tree/master/goemotions}{GoEmotions} \citep{demszky-etal-2020-goemotions} & Reddit comments with 27 emotion categories. & ``You're in for a treat- enjoy :)'' \\
  \arrayrulecolor{black!25}\specialrule{0.2pt}{2pt}{2pt}\arrayrulecolor{black}
  \raggedright\href{https://www.romanklinger.de/data-sets/}{GoodNewsEveryone} \citep{bostan-etal-2020-goodnewseveryone} & News headlines with emotions, roles and reader emotions. & ``Uncle Sam offering \$1K to people who adopt [...]'' \\
  \arrayrulecolor{black!25}\specialrule{0.2pt}{2pt}{2pt}\arrayrulecolor{black}
  \raggedright\href{http://implicitemotions.wassa2018.com/data}{IEST} \citep{klinger-etal-2018-iest} & Tweets labelled by an emotion word that was removed. & ``i'm so \dots{} because i bought a jar of pickles!!'' \\
  \arrayrulecolor{black!25}\specialrule{0.2pt}{2pt}{2pt}\arrayrulecolor{black}
  \raggedright\href{https://www.unige.ch/cisa/research/materials-and-online-research/research-material/}{ISEAR} \citep{scherer-wallbott-1994-evidence} & Events of students in 37 countries, own emotions. & ``I received a letter from a distant friend.'' \\
  \arrayrulecolor{black!25}\specialrule{0.2pt}{2pt}{2pt}\arrayrulecolor{black}
  \raggedright\href{https://github.com/declare-lab/MELD}{MELD} \citep{poria-etal-2019-meld} & Utterances from the TV series Friends, seven emotions. & ``You know, you shouldn't even be here, it's a school night.'' \\
  \arrayrulecolor{black!25}\specialrule{0.2pt}{2pt}{2pt}\arrayrulecolor{black}
  \raggedright\href{https://www.romanklinger.de/data-sets/}{MMEmo} \citep{khlyzova-etal-2022-complementarity} & Reddit posts with images, writers' emotions (text only). & ``This really made me smile!'' \\
  \arrayrulecolor{black!25}\specialrule{0.2pt}{2pt}{2pt}\arrayrulecolor{black}
  \raggedright\href{https://www.uni-bamberg.de/en/nlproc/projects/item/}{Mult\textsuperscript{2}EMo} \citep{bagdon-etal-2026-emotion} & Social media posts, authors' emotions and appraisals. & ``Feeling happy on this sunny dog walk'' \\
  \arrayrulecolor{black!25}\specialrule{0.2pt}{2pt}{2pt}\arrayrulecolor{black}
  \raggedright\href{https://www.romanklinger.de/data-sets/}{Nonsense words} \citep{sabbatino-etal-2022-splink} & Nonsense words with emotion intensities (a lexicon). & ``zel'' \\
  \arrayrulecolor{black!25}\specialrule{0.2pt}{2pt}{2pt}\arrayrulecolor{black}
  \raggedright\href{https://github.com/tnhaider/poetry-emotion}{PO-EMO} \citep{haider-etal-2020-po} & Lines of poems with aesthetic emotions (de, en). & ``Die Bächlein von den Bergen springen,'' \\
  \arrayrulecolor{black!25}\specialrule{0.2pt}{2pt}{2pt}\arrayrulecolor{black}
  \raggedright\href{https://www.romanklinger.de/data-sets/}{REMAN} \citep{kim-klinger-2018-feels} & Literature with emotions, cues, experiencers, targets, causes. & ``[\dots] Has that wicked little Blanche been making mischief?'' \\
  \arrayrulecolor{black!25}\specialrule{0.2pt}{2pt}{2pt}\arrayrulecolor{black}
  \raggedright\href{https://www.romanklinger.de/ssec/}{SSEC} \citep{schuff-etal-2017-annotation} & Tweets on the 2016 US election, Plutchik's emotions. & ``[\dots] I am ready to buy some Hillary gear'' \\
  \arrayrulecolor{black!25}\specialrule{0.2pt}{2pt}{2pt}\arrayrulecolor{black}
  \raggedright\href{https://people.rc.rit.edu/coagla/affectdata/index.html}{Tales} \citep{alm-2008-affect} & Sentences of fairy tales with high annotator agreement. & ``Then Tiny wept, and said she would not marry the disagreeable mole.'' \\
  \arrayrulecolor{black!25}\specialrule{0.2pt}{2pt}{2pt}\arrayrulecolor{black}
  \raggedright\href{https://saifmohammad.com/WebPages/SentimentEmotionLabeledData.html}{TEC} \citep{mohammad-2012-emotional} & Tweets labelled by the emotion hashtag of the writer. & ``Birthday and Christmas holidays coming up around the corner'' \\
  \arrayrulecolor{black!25}\specialrule{0.2pt}{2pt}{2pt}\arrayrulecolor{black}
  \raggedright\href{https://www.romanklinger.de/data-sets/}{x-enVENT} \citep{troiano-etal-2022-x} & enISEAR events, all experiencers' emotions, appraisals. & ``I felt \dots{} when my daughter was born.'' \\
  \arrayrulecolor{black!25}\specialrule{0.2pt}{2pt}{2pt}\arrayrulecolor{black}
  \raggedright\href{https://github.com/Helsinki-NLP/XED}{XED} \citep{ohman-etal-2020-xed} & Movie subtitles with Plutchik's emotions (en, de). & ``Don't look so depressed.'' \\
  \bottomrule
  \end{tabularx}
  \caption{Corpora in the \hub{} (48). The name links to the place where the data was originally published. Examples are shortened ([\dots]); `\dots' in the ISEAR-style texts and IEST is the masked emotion word.}\label{tab:corpora}
\end{table*}
 

\begin{thebibliography}{81}
\providecommand{\natexlab}[1]{#1}

\bibitem[{Acheampong et~al.(2020)Acheampong, Wenyu, and
  Nunoo-Mensah}]{acheampong-etal-2020-text}
Francisca~Adoma Acheampong, Chen Wenyu, and Henry Nunoo-Mensah. 2020.
\newblock \href {https://doi.org/10.1002/eng2.12189} {Text-based emotion
  detection: Advances, challenges, and opportunities}.
\newblock \emph{Engineering Reports}, 2(7):e12189.

\bibitem[{Alm(2008)}]{alm-2008-affect}
Cecilia~Ovesdotter Alm. 2008.
\newblock \href {https://people.rc.rit.edu/coagla/affectdata/index.html}
  {\emph{Affect in Text and Speech}}.
\newblock Ph.D. thesis, University of Illinois at Urbana-Champaign.

\bibitem[{Alm et~al.(2005)Alm, Roth, and Sproat}]{alm-etal-2005-emotions}
Cecilia~Ovesdotter Alm, Dan Roth, and Richard Sproat. 2005.
\newblock \href {https://aclanthology.org/H05-1073/} {Emotions from text:
  Machine learning for text-based emotion prediction}.
\newblock In \emph{Proceedings of Human Language Technology Conference and
  Conference on Empirical Methods in Natural Language Processing}, pages
  579--586, Vancouver, British Columbia, Canada. Association for Computational
  Linguistics.

\bibitem[{Alswaidan and Menai(2020)}]{alswaidan-menai-2020-survey}
Nourah Alswaidan and Mohamed El~Bachir Menai. 2020.
\newblock \href {https://doi.org/10.1007/s10115-020-01449-0} {A survey of
  state-of-the-art approaches for emotion recognition in text}.
\newblock \emph{Knowledge and Information Systems}, 62(8):2937--2987.

\bibitem[{Antypas et~al.(2023)Antypas, Ushio, Barbieri, Neves, Rezaee,
  Espinosa-Anke, Pei, and Camacho-Collados}]{antypas-etal-2023-supertweeteval}
Dimosthenis Antypas, Asahi Ushio, Francesco Barbieri, Leonardo Neves, Kiamehr
  Rezaee, Luis Espinosa-Anke, Jiaxin Pei, and Jose Camacho-Collados. 2023.
\newblock \href {https://doi.org/10.18653/v1/2023.findings-emnlp.838}
  {{S}uper{T}weet{E}val: A challenging, unified and heterogeneous benchmark for
  social media {NLP} research}.
\newblock In \emph{Findings of the {A}ssociation for {C}omputational
  {L}inguistics: {EMNLP} 2023}, pages 12590--12607, Singapore. Association for
  Computational Linguistics.

\bibitem[{Bagdon et~al.(2025)Bagdon, Combs, Silberer, and
  Klinger}]{bagdon-etal-2025-donate}
Christopher Bagdon, Aidan Combs, Carina Silberer, and Roman Klinger. 2025.
\newblock \href {https://doi.org/10.18653/v1/2025.acl-long.847} {Donate or
  create? comparing data collection strategies for emotion-labeled multimodal
  social media posts}.
\newblock In \emph{Proceedings of the 63rd Annual Meeting of the {A}ssociation
  for {C}omputational {L}inguistics (Volume 1: Long Papers)}, pages
  17307--17330, Vienna, Austria. Association for Computational Linguistics.

\bibitem[{Bagdon et~al.(2026)Bagdon, Silberer, and
  Klinger}]{bagdon-etal-2026-emotion}
Christopher Bagdon, Carina Silberer, and Roman Klinger. 2026.
\newblock \href {https://doi.org/10.18653/2026.emnlp-main.306} {Emotion
  experience, expression, and perception: Emotion analysis on multimodal social
  media posts}.
\newblock In \emph{Proceedings of the 2026 Conference on Empirical Methods in
  Natural Language Processing}, pages 7051--7078, Budapest, Hungary.
  Association for Computational Linguistics.

\bibitem[{Barbieri et~al.(2020)Barbieri, Camacho-Collados, Espinosa~Anke, and
  Neves}]{barbieri-etal-2020-tweeteval}
Francesco Barbieri, Jose Camacho-Collados, Luis Espinosa~Anke, and Leonardo
  Neves. 2020.
\newblock \href {https://doi.org/10.18653/v1/2020.findings-emnlp.148}
  {{T}weet{E}val: Unified benchmark and comparative evaluation for tweet
  classification}.
\newblock In \emph{Findings of the {A}ssociation for {C}omputational
  {L}inguistics: {EMNLP} 2020}, pages 1644--1650, Online. Association for
  Computational Linguistics.

\bibitem[{Belay et~al.(2025)Belay, Azime, Ayele, Sidorov, Klakow, Slusallek,
  Kolesnikova, and Yimam}]{belay-etal-2025-evaluating}
Tadesse~Destaw Belay, Israel~Abebe Azime, Abinew~Ali Ayele, Grigori Sidorov,
  Dietrich Klakow, Philip Slusallek, Olga Kolesnikova, and Seid~Muhie Yimam.
  2025.
\newblock \href {https://aclanthology.org/2025.coling-main.237/} {Evaluating
  the capabilities of large language models for multi-label emotion
  understanding}.
\newblock In \emph{Proceedings of the 31st International Conference on
  Computational Linguistics}, pages 3523--3540, Abu Dhabi, UAE. Association for
  Computational Linguistics.

\bibitem[{Bostan et~al.(2020)Bostan, Kim, and
  Klinger}]{bostan-etal-2020-goodnewseveryone}
Laura Ana~Maria Bostan, Evgeny Kim, and Roman Klinger. 2020.
\newblock \href {https://aclanthology.org/2020.lrec-1.194/}
  {{G}ood{N}ews{E}veryone: A corpus of news headlines annotated with emotions,
  semantic roles, and reader perception}.
\newblock In \emph{Proceedings of the Twelfth Language Resources and Evaluation
  Conference}, pages 1554--1566, Marseille, France. European Language Resources
  Association.

\bibitem[{Bostan and Klinger(2018)}]{bostan-klinger-2018-analysis}
Laura-Ana-Maria Bostan and Roman Klinger. 2018.
\newblock \href {https://aclanthology.org/C18-1179/} {An analysis of annotated
  corpora for emotion classification in text}.
\newblock In \emph{Proceedings of the 27th International Conference on
  Computational Linguistics}, pages 2104--2119, Santa Fe, New Mexico, USA.
  Association for Computational Linguistics.

\bibitem[{Buechel and Hahn(2017)}]{buechel-hahn-2017-emobank}
Sven Buechel and Udo Hahn. 2017.
\newblock \href {https://aclanthology.org/E17-2092/} {{E}mo{B}ank: Studying the
  impact of annotation perspective and representation format on dimensional
  emotion analysis}.
\newblock In \emph{Proceedings of the 15th Conference of the {E}uropean Chapter
  of the {A}ssociation for {C}omputational {L}inguistics: Volume 2, Short
  Papers}, pages 578--585, Valencia, Spain. Association for Computational
  Linguistics.

\bibitem[{Camacho-Collados et~al.(2022)Camacho-Collados, Rezaee, Riahi, Ushio,
  Loureiro, Antypas, Boisson, Espinosa-Anke, Liu, Mart{\'i}nez-C{\'a}mara,
  Medina, Buhrmann, Neves, and Barbieri}]{camacho-collados-etal-2022-tweetnlp}
Jose Camacho-Collados, Kiamehr Rezaee, Talayeh Riahi, Asahi Ushio, Daniel
  Loureiro, Dimosthenis Antypas, Joanne Boisson, Luis Espinosa-Anke, Fangyu
  Liu, Eugenio Mart{\'i}nez-C{\'a}mara, Gonzalo Medina, Thomas Buhrmann,
  Leonardo Neves, and Francesco Barbieri. 2022.
\newblock \href {https://doi.org/10.18653/v1/2022.emnlp-demos.5} {{T}weet{NLP}:
  Cutting-edge natural language processing for social media}.
\newblock In \emph{Proceedings of the 2022 Conference on Empirical Methods in
  Natural Language Processing: System Demonstrations}, pages 38--49, Abu Dhabi,
  UAE. Association for Computational Linguistics.

\bibitem[{Campagnano et~al.(2022)Campagnano, Conia, and
  Navigli}]{campagnano-etal-2022-srl4e}
Cesare Campagnano, Simone Conia, and Roberto Navigli. 2022.
\newblock \href {https://doi.org/10.18653/v1/2022.acl-long.314} {{SRL4E} {--}
  {S}emantic {R}ole {L}abeling for {E}motions: {A} unified evaluation
  framework}.
\newblock In \emph{Proceedings of the 60th Annual Meeting of the {A}ssociation
  for {C}omputational {L}inguistics (Volume 1: Long Papers)}, pages 4586--4601,
  Dublin, Ireland. Association for Computational Linguistics.

\bibitem[{Canales and
  Mart{\'i}nez-Barco(2014)}]{canales-martinez-barco-2014-emotion}
Lea Canales and Patricio Mart{\'i}nez-Barco. 2014.
\newblock \href {https://doi.org/10.3115/v1/W14-6905} {Emotion detection from
  text: A survey}.
\newblock In \emph{Proceedings of the Workshop on Natural Language Processing
  in the 5th Information Systems Research Working Days ({JISIC})}, pages
  37--43, Quito, Ecuador. Association for Computational Linguistics.

\bibitem[{Casel et~al.(2021)Casel, Heindl, and
  Klinger}]{casel-etal-2021-emotion}
Felix Casel, Amelie Heindl, and Roman Klinger. 2021.
\newblock \href {https://aclanthology.org/2021.konvens-1.5/} {Emotion
  recognition under consideration of the emotion component process model}.
\newblock In \emph{Proceedings of the 17th Conference on Natural Language
  Processing ({KONVENS} 2021)}, pages 49--61, D{\"u}sseldorf, Germany. KONVENS
  2021 Organizers.

\bibitem[{{CrowdFlower}(2016)}]{crowdflower-2016-emotion}
{CrowdFlower}. 2016.
\newblock The emotion in text.
\newblock \url{https://data.world/crowdflower/sentiment-analysis-in-text}.

\bibitem[{Demszky et~al.(2020)Demszky, Movshovitz-Attias, Ko, Cowen, Nemade,
  and Ravi}]{demszky-etal-2020-goemotions}
Dorottya Demszky, Dana Movshovitz-Attias, Jeongwoo Ko, Alan Cowen, Gaurav
  Nemade, and Sujith Ravi. 2020.
\newblock \href {https://doi.org/10.18653/v1/2020.acl-main.372}
  {{G}o{E}motions: A dataset of fine-grained emotions}.
\newblock In \emph{Proceedings of the 58th Annual Meeting of the {A}ssociation
  for {C}omputational {L}inguistics}, pages 4040--4054, Online. Association for
  Computational Linguistics.

\bibitem[{Deng and Ren(2023)}]{deng-ren-2023-survey}
Jiawen Deng and Fuji Ren. 2023.
\newblock \href {https://doi.org/10.1109/TAFFC.2021.3053275} {A survey of
  textual emotion recognition and its challenges}.
\newblock \emph{IEEE Transactions on Affective Computing}, 14(1):49--67.

\bibitem[{Doan~Dang et~al.(2021)Doan~Dang, Oberl{\"a}nder, and
  Klinger}]{doan-dang-etal-2021-emotion}
Bao~Minh Doan~Dang, Laura Oberl{\"a}nder, and Roman Klinger. 2021.
\newblock \href {https://aclanthology.org/2021.konvens-1.7/} {Emotion stimulus
  detection in {G}erman news headlines}.
\newblock In \emph{Proceedings of the 17th Conference on Natural Language
  Processing ({KONVENS} 2021)}, pages 73--85, D{\"u}sseldorf, Germany. KONVENS
  2021 Organizers.

\bibitem[{El-Haj and Takanami(2023)}]{el-haj-takanami-2023-unifying}
Mo~El-Haj and Ryutaro Takanami. 2023.
\newblock \href {https://aclanthology.org/2023.ldk-1.19/} {Unifying emotion
  analysis datasets using valence arousal dominance ({VAD})}.
\newblock In \emph{Proceedings of the 4th Conference on Language, Data and
  Knowledge}, pages 220--225, Vienna, Austria. NOVA CLUNL, Portugal.

\bibitem[{Francisco et~al.(2012)Francisco, Herv{\'a}s, Peinado, and
  Gerv{\'a}s}]{francisco-etal-2012-emotales}
Virginia Francisco, Raquel Herv{\'a}s, Federico Peinado, and Pablo Gerv{\'a}s.
  2012.
\newblock \href {https://doi.org/10.1007/s10579-011-9140-5} {{EmoTales}:
  Creating a corpus of folk tales with emotional annotations}.
\newblock \emph{Language Resources and Evaluation}, 46(3):341--381.

\bibitem[{Ghazi et~al.(2015)Ghazi, Inkpen, and
  Szpakowicz}]{ghazi-etal-2015-detecting}
Diman Ghazi, Diana Inkpen, and Stan Szpakowicz. 2015.
\newblock \href {https://doi.org/10.1007/978-3-319-18117-2_12} {Detecting
  emotion stimuli in emotion-bearing sentences}.
\newblock In \emph{Computational Linguistics and Intelligent Text Processing
  (CICLing 2015)}, volume 9042 of \emph{Lecture Notes in Computer Science},
  pages 152--165. Springer.

\bibitem[{Gr{\"a}{\ss}er et~al.(2018)Gr{\"a}{\ss}er, Kallumadi, Malberg, and
  Zaunseder}]{grasser-etal-2018-aspect}
Felix Gr{\"a}{\ss}er, Surya Kallumadi, Hagen Malberg, and Sebastian Zaunseder.
  2018.
\newblock \href {https://doi.org/10.1145/3194658.3194677} {Aspect-based
  sentiment analysis of drug reviews applying cross-domain and cross-data
  learning}.
\newblock In \emph{Proceedings of the 2018 International Conference on Digital
  Health}, pages 121--125. ACM.

\bibitem[{Greschner et~al.(2026)Greschner, Weber, and
  Klinger}]{greschner-etal-2026-trust}
Lynn Greschner, Sabine Weber, and Roman Klinger. 2026.
\newblock \href {https://doi.org/10.63317/484rpnvebop5} {Trust me, {I} can
  convince you: The contextualized argument appraisal framework and the
  {C}ont{A}rg{A} corpus}.
\newblock In \emph{Proceedings of the Fifteenth Language Resources and
  Evaluation Conference}, pages 8327--8346, Palma de Mallorca, Spain. ELRA
  Language Resource Association.

\bibitem[{Haider et~al.(2020)Haider, Eger, Kim, Klinger, and
  Menninghaus}]{haider-etal-2020-po}
Thomas Haider, Steffen Eger, Evgeny Kim, Roman Klinger, and Winfried
  Menninghaus. 2020.
\newblock \href {https://aclanthology.org/2020.lrec-1.205/} {{PO}-{EMO}:
  Conceptualization, annotation, and modeling of aesthetic emotions in {G}erman
  and {E}nglish poetry}.
\newblock In \emph{Proceedings of the Twelfth Language Resources and Evaluation
  Conference}, pages 1652--1663, Marseille, France. European Language Resources
  Association.

\bibitem[{Hartmann(2022)}]{hartmann-2022-emotion}
Jochen Hartmann. 2022.
\newblock Emotion {English} {DistilRoBERTa}-base.
\newblock
  \url{https://huggingface.co/j-hartmann/emotion-english-distilroberta-base/}.

\bibitem[{Hegde and Jayalath(2025)}]{hegde-jayalath-2025-emotions}
Karishma Hegde and Hemadri Jayalath. 2025.
\newblock \href {https://arxiv.org/abs/2505.01542} {Emotions in the loop: A
  survey of affective computing for emotional support}.
\newblock \emph{Preprint}, arXiv:2505.01542.

\bibitem[{Hofmann et~al.(2021)Hofmann, Troiano, and
  Klinger}]{hofmann-etal-2021-emotion}
Jan Hofmann, Enrica Troiano, and Roman Klinger. 2021.
\newblock \href {https://aclanthology.org/2021.wassa-1.17/} {Emotion-aware,
  emotion-agnostic, or automatic: Corpus creation strategies to obtain
  cognitive event appraisal annotations}.
\newblock In \emph{Proceedings of the Eleventh Workshop on Computational
  Approaches to Subjectivity, Sentiment and Social Media Analysis}, pages
  160--170, Online. Association for Computational Linguistics.

\bibitem[{Hofmann et~al.(2020)Hofmann, Troiano, Sassenberg, and
  Klinger}]{hofmann-etal-2020-appraisal}
Jan Hofmann, Enrica Troiano, Kai Sassenberg, and Roman Klinger. 2020.
\newblock \href {https://doi.org/10.18653/v1/2020.coling-main.11} {Appraisal
  theories for emotion classification in text}.
\newblock In \emph{Proceedings of the 28th International Conference on
  Computational Linguistics}, pages 125--138, Barcelona, Spain (Online).
  International Committee on Computational Linguistics.

\bibitem[{Khlyzova et~al.(2022)Khlyzova, Silberer, and
  Klinger}]{khlyzova-etal-2022-complementarity}
Anna Khlyzova, Carina Silberer, and Roman Klinger. 2022.
\newblock \href {https://doi.org/10.18653/v1/2022.wassa-1.1} {On the
  complementarity of images and text for the expression of emotions in social
  media}.
\newblock In \emph{Proceedings of the 12th Workshop on Computational Approaches
  to Subjectivity, Sentiment {\&} Social Media Analysis}, pages 1--15, Dublin,
  Ireland. Association for Computational Linguistics.

\bibitem[{Kim and Klinger(2018)}]{kim-klinger-2018-feels}
Evgeny Kim and Roman Klinger. 2018.
\newblock \href {https://aclanthology.org/C18-1114/} {Who feels what and why?
  annotation of a literature corpus with semantic roles of emotions}.
\newblock In \emph{Proceedings of the 27th International Conference on
  Computational Linguistics}, pages 1345--1359, Santa Fe, New Mexico, USA.
  Association for Computational Linguistics.

\bibitem[{Kim and Klinger(2019{\natexlab{a}})}]{kim-klinger-2019-analysis}
Evgeny Kim and Roman Klinger. 2019{\natexlab{a}}.
\newblock \href {https://doi.org/10.18653/v1/W19-3406} {An analysis of emotion
  communication channels in fan-fiction: Towards emotional storytelling}.
\newblock In \emph{Proceedings of the Second Workshop on Storytelling}, pages
  56--64, Florence, Italy. Association for Computational Linguistics.

\bibitem[{Kim and Klinger(2019{\natexlab{b}})}]{kim-klinger-2019-frowning}
Evgeny Kim and Roman Klinger. 2019{\natexlab{b}}.
\newblock \href {https://doi.org/10.18653/v1/N19-1067} {Frowning {F}rodo,
  wincing {L}eia, and a seriously great friendship: Learning to classify
  emotional relationships of fictional characters}.
\newblock In \emph{Proceedings of the 2019 Conference of the {N}orth {A}merican
  Chapter of the {A}ssociation for {C}omputational {L}inguistics: Human
  Language Technologies, Volume 1 (Long and Short Papers)}, pages 647--653,
  Minneapolis, Minnesota. Association for Computational Linguistics.

\bibitem[{Kim and Klinger(2019{\natexlab{c}})}]{kim-klinger-2019-survey}
Evgeny Kim and Roman Klinger. 2019{\natexlab{c}}.
\newblock \href {https://doi.org/10.17175/2019_008} {A survey on sentiment and
  emotion analysis for computational literary studies}.
\newblock \emph{Zeitschrift f{\"u}r digitale Geisteswissenschaften}, 4.

\bibitem[{Klinger(2023)}]{klinger-2023-event}
Roman Klinger. 2023.
\newblock \href {https://doi.org/10.18653/v1/2023.bigpicture-1.1} {Where are we
  in event-centric emotion analysis? bridging emotion role labeling and
  appraisal-based approaches}.
\newblock In \emph{Proceedings of the Big Picture Workshop}, pages 1--17,
  Singapore. Association for Computational Linguistics.

\bibitem[{Klinger et~al.(2018)Klinger, De~Clercq, Mohammad, and
  Balahur}]{klinger-etal-2018-iest}
Roman Klinger, Orph{\'e}e De~Clercq, Saif Mohammad, and Alexandra Balahur.
  2018.
\newblock \href {https://doi.org/10.18653/v1/W18-6206} {{IEST}: {WASSA}-2018
  implicit emotions shared task}.
\newblock In \emph{Proceedings of the 9th Workshop on Computational Approaches
  to Subjectivity, Sentiment and Social Media Analysis}, pages 31--42,
  Brussels, Belgium. Association for Computational Linguistics.

\bibitem[{Koufakou et~al.(2024)Koufakou, Nieves, and
  Peller}]{koufakou-etal-2024-towards}
Anna Koufakou, Elijah Nieves, and John Peller. 2024.
\newblock \href {https://doi.org/10.18653/v1/2024.genbench-1.13} {Towards a new
  benchmark for emotion detection in {NLP}: A unifying framework of recent
  corpora}.
\newblock In \emph{Proceedings of the 2nd {G}en{B}ench Workshop on
  Generalisation (Benchmarking) in {NLP}}, pages 196--206, Miami, Florida, USA.
  Association for Computational Linguistics.

\bibitem[{Kulkarni(2018)}]{kulkarni-2018-million}
Rohit Kulkarni. 2018.
\newblock \href {https://doi.org/10.7910/DVN/SYBGZL} {A million news
  headlines}.

\bibitem[{Li et~al.(2017)Li, Su, Shen, Li, Cao, and
  Niu}]{li-etal-2017-dailydialog}
Yanran Li, Hui Su, Xiaoyu Shen, Wenjie Li, Ziqiang Cao, and Shuzi Niu. 2017.
\newblock \href {https://aclanthology.org/I17-1099/} {{D}aily{D}ialog: A
  manually labelled multi-turn dialogue dataset}.
\newblock In \emph{Proceedings of the Eighth International Joint Conference on
  Natural Language Processing (Volume 1: Long Papers)}, pages 986--995, Taipei,
  Taiwan. Asian Federation of Natural Language Processing.

\bibitem[{Liu et~al.(2024)Liu, Yang, Xie, Zhang, and
  Ananiadou}]{liu-etal-2024-emollms}
Zhiwei Liu, Kailai Yang, Qianqian Xie, Tianlin Zhang, and Sophia Ananiadou.
  2024.
\newblock \href {https://doi.org/10.1145/3637528.3671552} {{EmoLLMs}: A series
  of emotional large language models and annotation tools for comprehensive
  affective analysis}.
\newblock In \emph{Proceedings of the 30th ACM SIGKDD Conference on Knowledge
  Discovery and Data Mining}, pages 5487--5496.

\bibitem[{Low et~al.(2020)Low, Rumker, Talkar, Torous, Cecchi, and
  Ghosh}]{low-etal-2020-natural}
Daniel~M. Low, Laurie Rumker, Tanya Talkar, John Torous, Guillermo Cecchi, and
  Satrajit~S. Ghosh. 2020.
\newblock \href {https://doi.org/10.2196/22635} {Natural language processing
  reveals vulnerable mental health support groups and heightened health anxiety
  on {R}eddit during {COVID}-19: Observational study}.
\newblock \emph{Journal of Medical Internet Research}, 22(10):e22635.

\bibitem[{Misra(2022)}]{misra-2022-news}
Rishabh Misra. 2022.
\newblock \href {https://arxiv.org/abs/2209.11429} {News category dataset}.
\newblock \emph{Preprint}, arXiv:2209.11429.

\bibitem[{Mohammad(2012)}]{mohammad-2012-emotional}
Saif Mohammad. 2012.
\newblock \href {https://aclanthology.org/S12-1033/} {{\#}emotional tweets}.
\newblock In \emph{*{SEM} 2012: The First Joint Conference on Lexical and
  Computational Semantics {--} Volume 1: Proceedings of the main conference and
  the shared task, and Volume 2: Proceedings of the Sixth {I}nternational
  {W}orkshop on {S}emantic {E}valuation ({S}em{E}val 2012)}, pages 246--255,
  Montr{\'e}al, Canada. Association for Computational Linguistics.

\bibitem[{Mohammad(2023)}]{mohammad-2023-best}
Saif Mohammad. 2023.
\newblock \href {https://doi.org/10.18653/v1/2023.findings-eacl.136} {Best
  practices in the creation and use of emotion lexicons}.
\newblock In \emph{Findings of the {A}ssociation for {C}omputational
  {L}inguistics: {EACL} 2023}, pages 1825--1836, Dubrovnik, Croatia.
  Association for Computational Linguistics.

\bibitem[{Mohammad and Bravo-Marquez(2017)}]{mohammad-bravo-marquez-2017-wassa}
Saif Mohammad and Felipe Bravo-Marquez. 2017.
\newblock \href {https://doi.org/10.18653/v1/W17-5205} {{WASSA}-2017 shared
  task on emotion intensity}.
\newblock In \emph{Proceedings of the 8th Workshop on Computational Approaches
  to Subjectivity, Sentiment and Social Media Analysis}, pages 34--49,
  Copenhagen, Denmark. Association for Computational Linguistics.

\bibitem[{Mohammad et~al.(2018)Mohammad, Bravo-Marquez, Salameh, and
  Kiritchenko}]{mohammad-etal-2018-semeval}
Saif Mohammad, Felipe Bravo-Marquez, Mohammad Salameh, and Svetlana
  Kiritchenko. 2018.
\newblock \href {https://doi.org/10.18653/v1/S18-1001} {{S}em{E}val-2018 task
  1: Affect in tweets}.
\newblock In \emph{Proceedings of the 12th {I}nternational {W}orkshop on
  {S}emantic {E}valuation}, pages 1--17, New Orleans, Louisiana. Association
  for Computational Linguistics.

\bibitem[{Mohammad et~al.(2014)Mohammad, Zhu, and
  Martin}]{mohammad-etal-2014-semantic}
Saif Mohammad, Xiaodan Zhu, and Joel Martin. 2014.
\newblock \href {https://doi.org/10.3115/v1/W14-2607} {Semantic role labeling
  of emotions in tweets}.
\newblock In \emph{Proceedings of the 5th Workshop on Computational Approaches
  to Subjectivity, Sentiment and Social Media Analysis}, pages 32--41,
  Baltimore, Maryland. Association for Computational Linguistics.

\bibitem[{Mohammad(2016)}]{mohammad-2016-sentiment}
Saif~M. Mohammad. 2016.
\newblock \href {https://doi.org/10.1016/B978-0-08-100508-8.00009-6} {Sentiment
  analysis: Detecting valence, emotions, and other affectual states from text}.
\newblock In Herbert~L. Meiselman, editor, \emph{Emotion Measurement}, pages
  201--237. Woodhead Publishing.

\bibitem[{Mohammad(2022)}]{mohammad-2022-ethics-sheet}
Saif~M. Mohammad. 2022.
\newblock \href {https://doi.org/10.1162/coli_a_00433} {Ethics sheet for
  automatic emotion recognition and sentiment analysis}.
\newblock \emph{Computational Linguistics}, 48(2):239--278.

\bibitem[{Mohammad et~al.(2015)Mohammad, Zhu, Kiritchenko, and
  Martin}]{mohammad-etal-2015-electoral}
Saif~M. Mohammad, Xiaodan Zhu, Svetlana Kiritchenko, and Joel Martin. 2015.
\newblock \href {https://doi.org/10.1016/j.ipm.2014.09.003} {Sentiment,
  emotion, purpose, and style in electoral tweets}.
\newblock \emph{Information Processing \& Management}, 51(4):480--499.

\bibitem[{Muhammad et~al.(2025)Muhammad, Ousidhoum, Abdulmumin, Wahle, Ruas,
  Beloucif, de~Kock, Surange, Teodorescu, Ahmad, Adelani, Aji, Ali, Alimova,
  Araujo, Babakov, Baes, Bucur, Bukula, Cao, Tufi{\~n}o, Chevi, Chukwuneke,
  Ciobotaru, Dementieva, Gadanya, Geislinger, Gipp, Hourrane, Ignat, Lawan,
  Mabuya, Mahendra, Marivate, Panchenko, Piper, Ferreira, Protasov, Rutunda,
  Shrivastava, Udrea, Wanzare, Wu, Wunderlich, Zhafran, Zhang, Zhou, and
  Mohammad}]{muhammad-etal-2025-brighter}
Shamsuddeen~Hassan Muhammad, Nedjma Ousidhoum, Idris Abdulmumin, Jan~Philip
  Wahle, Terry Ruas, Meriem Beloucif, Christine de~Kock, Nirmal Surange,
  Daniela Teodorescu, Ibrahim~Said Ahmad, David~Ifeoluwa Adelani, Alham~Fikri
  Aji, Felermino D. M.~A. Ali, Ilseyar Alimova, Vladimir Araujo, Nikolay
  Babakov, Naomi Baes, Ana-Maria Bucur, Andiswa Bukula, and 29 others. 2025.
\newblock \href {https://doi.org/10.18653/v1/2025.acl-long.436} {{BRIGHTER}:
  {BRI}dging the gap in human-annotated textual emotion recognition datasets
  for 28 languages}.
\newblock In \emph{Proceedings of the 63rd Annual Meeting of the {A}ssociation
  for {C}omputational {L}inguistics (Volume 1: Long Papers)}, pages 8895--8916,
  Vienna, Austria. Association for Computational Linguistics.

\bibitem[{{\"O}hman et~al.(2020){\"O}hman, P{\`a}mies, Kajava, and
  Tiedemann}]{ohman-etal-2020-xed}
Emily {\"O}hman, Marc P{\`a}mies, Kaisla Kajava, and J{\"o}rg Tiedemann. 2020.
\newblock \href {https://doi.org/10.18653/v1/2020.coling-main.575} {{XED}: A
  multilingual dataset for sentiment analysis and emotion detection}.
\newblock In \emph{Proceedings of the 28th International Conference on
  Computational Linguistics}, pages 6542--6552, Barcelona, Spain (Online).
  International Committee on Computational Linguistics.

\bibitem[{P{\'e}rez et~al.(2021)P{\'e}rez, Rajngewerc, Giudici, Furman, Luque,
  Alemany, and Mart{\'i}nez}]{perez-etal-2021-pysentimiento}
Juan~Manuel P{\'e}rez, Mariela Rajngewerc, Juan~Carlos Giudici, Dami{\'a}n~A.
  Furman, Franco Luque, Laura~Alonso Alemany, and Mar{\'i}a~Vanina
  Mart{\'i}nez. 2021.
\newblock \href {https://arxiv.org/abs/2106.09462} {pysentimiento: A {Python}
  toolkit for opinion mining and social {NLP} tasks}.
\newblock \emph{Preprint}, arXiv:2106.09462.

\bibitem[{Plaza-del Arco et~al.(2024)Plaza-del Arco, Cercas~Curry,
  Cercas~Curry, and Hovy}]{plaza-del-arco-etal-2024-emotion}
Flor~Miriam Plaza-del Arco, Alba~A. Cercas~Curry, Amanda Cercas~Curry, and Dirk
  Hovy. 2024.
\newblock \href {https://aclanthology.org/2024.lrec-main.506/} {Emotion
  analysis in {NLP}: Trends, gaps and roadmap for future directions}.
\newblock In \emph{Proceedings of the 2024 Joint International Conference on
  Computational Linguistics, Language Resources and Evaluation ({LREC}-{COLING}
  2024)}, pages 5696--5710, Torino, Italia. ELRA and ICCL.

\bibitem[{Plaza~del Arco et~al.(2020)Plaza~del Arco, Strapparava, Urena~Lopez,
  and Martin}]{plaza-del-arco-etal-2020-emoevent}
Flor~Miriam Plaza~del Arco, Carlo Strapparava, L.~Alfonso Urena~Lopez, and
  Maite Martin. 2020.
\newblock \href {https://aclanthology.org/2020.lrec-1.186/} {{E}mo{E}vent: A
  multilingual emotion corpus based on different events}.
\newblock In \emph{Proceedings of the Twelfth Language Resources and Evaluation
  Conference}, pages 1492--1498, Marseille, France. European Language Resources
  Association.

\bibitem[{Poria et~al.(2019)Poria, Hazarika, Majumder, Naik, Cambria, and
  Mihalcea}]{poria-etal-2019-meld}
Soujanya Poria, Devamanyu Hazarika, Navonil Majumder, Gautam Naik, Erik
  Cambria, and Rada Mihalcea. 2019.
\newblock \href {https://doi.org/10.18653/v1/P19-1050} {{MELD}: A multimodal
  multi-party dataset for emotion recognition in conversations}.
\newblock In \emph{Proceedings of the 57th Annual Meeting of the {A}ssociation
  for {C}omputational {L}inguistics}, pages 527--536, Florence, Italy.
  Association for Computational Linguistics.

\bibitem[{Preo{\c{t}}iuc-Pietro et~al.(2016)Preo{\c{t}}iuc-Pietro, Schwartz,
  Park, Eichstaedt, Kern, Ungar, and
  Shulman}]{preotiuc-pietro-etal-2016-modelling}
Daniel Preo{\c{t}}iuc-Pietro, H.~Andrew Schwartz, Gregory Park, Johannes
  Eichstaedt, Margaret Kern, Lyle Ungar, and Elisabeth Shulman. 2016.
\newblock \href {https://doi.org/10.18653/v1/W16-0404} {Modelling valence and
  arousal in {F}acebook posts}.
\newblock In \emph{Proceedings of the 7th Workshop on Computational Approaches
  to Subjectivity, Sentiment and Social Media Analysis}, pages 9--15, San
  Diego, California. Association for Computational Linguistics.

\bibitem[{Rashkin et~al.(2019)Rashkin, Smith, Li, and
  Boureau}]{rashkin-etal-2019-towards}
Hannah Rashkin, Eric~Michael Smith, Margaret Li, and Y-Lan Boureau. 2019.
\newblock \href {https://doi.org/10.18653/v1/P19-1534} {Towards empathetic
  open-domain conversation models: A new benchmark and dataset}.
\newblock In \emph{Proceedings of the 57th Annual Meeting of the {A}ssociation
  for {C}omputational {L}inguistics}, pages 5370--5381, Florence, Italy.
  Association for Computational Linguistics.

\bibitem[{Sabbatino et~al.(2022)Sabbatino, Troiano, Schweitzer, and
  Klinger}]{sabbatino-etal-2022-splink}
Valentino Sabbatino, Enrica Troiano, Antje Schweitzer, and Roman Klinger. 2022.
\newblock \href {https://doi.org/10.18653/v1/2022.wassa-1.4} {``splink'' is
  happy and ``phrouth'' is scary: Emotion intensity analysis for nonsense
  words}.
\newblock In \emph{Proceedings of the 12th Workshop on Computational Approaches
  to Subjectivity, Sentiment {\&} Social Media Analysis}, pages 37--50, Dublin,
  Ireland. Association for Computational Linguistics.

\bibitem[{Sabour et~al.(2024)Sabour, Liu, Zhang, Liu, Zhou, Sunaryo, Lee,
  Mihalcea, and Huang}]{sabour-etal-2024-emobench}
Sahand Sabour, Siyang Liu, Zheyuan Zhang, June Liu, Jinfeng Zhou, Alvionna
  Sunaryo, Tatia Lee, Rada Mihalcea, and Minlie Huang. 2024.
\newblock \href {https://doi.org/10.18653/v1/2024.acl-long.326} {{E}mo{B}ench:
  Evaluating the emotional intelligence of large language models}.
\newblock In \emph{Proceedings of the 62nd Annual Meeting of the {A}ssociation
  for {C}omputational {L}inguistics (Volume 1: Long Papers)}, pages 5986--6004,
  Bangkok, Thailand. Association for Computational Linguistics.

\bibitem[{Saravia et~al.(2018)Saravia, Liu, Huang, Wu, and
  Chen}]{saravia-etal-2018-carer}
Elvis Saravia, Hsien-Chi~Toby Liu, Yen-Hao Huang, Junlin Wu, and Yi-Shin Chen.
  2018.
\newblock \href {https://doi.org/10.18653/v1/D18-1404} {{CARER}: Contextualized
  affect representations for emotion recognition}.
\newblock In \emph{Proceedings of the 2018 Conference on Empirical Methods in
  Natural Language Processing}, pages 3687--3697, Brussels, Belgium.
  Association for Computational Linguistics.

\bibitem[{Scherer and Wallbott(1994)}]{scherer-wallbott-1994-evidence}
Klaus~R. Scherer and Harald~G. Wallbott. 1994.
\newblock \href {https://doi.org/10.1037/0022-3514.66.2.310} {Evidence for
  universality and cultural variation of differential emotion response
  patterning}.
\newblock \emph{Journal of Personality and Social Psychology}, 66(2):310--328.

\bibitem[{Schuff et~al.(2017)Schuff, Barnes, Mohme, Pad{\'o}, and
  Klinger}]{schuff-etal-2017-annotation}
Hendrik Schuff, Jeremy Barnes, Julian Mohme, Sebastian Pad{\'o}, and Roman
  Klinger. 2017.
\newblock \href {https://doi.org/10.18653/v1/W17-5203} {Annotation, modelling
  and analysis of fine-grained emotions on a stance and sentiment detection
  corpus}.
\newblock In \emph{Proceedings of the 8th Workshop on Computational Approaches
  to Subjectivity, Sentiment and Social Media Analysis}, pages 13--23,
  Copenhagen, Denmark. Association for Computational Linguistics.

\bibitem[{Song et~al.(2025)Song, Zhang, Gao, Yao, and
  Zhang}]{song-etal-2025-large}
Changhao Song, Yazhou Zhang, Hui Gao, Ben Yao, and Peng Zhang. 2025.
\newblock \href {https://arxiv.org/abs/2508.07959} {Large language models for
  subjective language understanding: A survey}.
\newblock \emph{Preprint}, arXiv:2508.07959.

\bibitem[{Stajner and Klinger(2023)}]{stajner-klinger-2023-emotion}
Sanja Stajner and Roman Klinger. 2023.
\newblock \href {https://doi.org/10.18653/v1/2023.eacl-tutorials.2} {Emotion
  analysis from texts}.
\newblock In \emph{Proceedings of the 17th Conference of the {E}uropean Chapter
  of the {A}ssociation for {C}omputational {L}inguistics: Tutorial Abstracts},
  pages 7--12, Dubrovnik, Croatia. Association for Computational Linguistics.

\bibitem[{Stranisci et~al.(2022)Stranisci, Frenda, Ceccaldi, Basile, Damiano,
  and Patti}]{stranisci-etal-2022-appreddit}
Marco~Antonio Stranisci, Simona Frenda, Eleonora Ceccaldi, Valerio Basile,
  Rossana Damiano, and Viviana Patti. 2022.
\newblock \href {https://aclanthology.org/2022.lrec-1.406/} {{APPR}eddit: a
  corpus of {R}eddit posts annotated for appraisal}.
\newblock In \emph{Proceedings of the Thirteenth Language Resources and
  Evaluation Conference}, pages 3809--3818, Marseille, France. European
  Language Resources Association.

\bibitem[{Strapparava and Mihalcea(2007)}]{strapparava-mihalcea-2007-semeval}
Carlo Strapparava and Rada Mihalcea. 2007.
\newblock \href {https://aclanthology.org/S07-1013/} {{S}em{E}val-2007 task 14:
  Affective text}.
\newblock In \emph{Proceedings of the Fourth International Workshop on Semantic
  Evaluations ({S}em{E}val-2007)}, pages 70--74, Prague, Czech Republic.
  Association for Computational Linguistics.

\bibitem[{Troiano et~al.(2024)Troiano, Labat, Stranisci, Damiano, Patti, and
  Klinger}]{troiano-etal-2024-dealing}
Enrica Troiano, Sofie Labat, Marco~Antonio Stranisci, Rossana Damiano, Viviana
  Patti, and Roman Klinger. 2024.
\newblock \href {https://doi.org/10.18653/v1/2024.findings-emnlp.89} {Dealing
  with controversy: An emotion and coping strategy corpus based on role
  playing}.
\newblock In \emph{Findings of the {A}ssociation for {C}omputational
  {L}inguistics: {EMNLP} 2024}, pages 1634--1658, Miami, Florida, USA.
  Association for Computational Linguistics.

\bibitem[{Troiano et~al.(2023)Troiano, Oberl{\"a}nder, and
  Klinger}]{troiano-etal-2023-dimensional}
Enrica Troiano, Laura Oberl{\"a}nder, and Roman Klinger. 2023.
\newblock \href {https://doi.org/10.1162/coli_a_00461} {Dimensional modeling of
  emotions in text with appraisal theories: Corpus creation, annotation
  reliability, and prediction}.
\newblock \emph{Computational Linguistics}, 49(1):1--72.

\bibitem[{Troiano et~al.(2022)Troiano, Oberl{\"a}nder, Wegge, and
  Klinger}]{troiano-etal-2022-x}
Enrica Troiano, Laura Oberl{\"a}nder, Maximilian Wegge, and Roman Klinger.
  2022.
\newblock \href {https://aclanthology.org/2022.lrec-1.146/} {x-en{VENT}: A
  corpus of event descriptions with experiencer-specific emotion and appraisal
  annotations}.
\newblock In \emph{Proceedings of the Thirteenth Language Resources and
  Evaluation Conference}, pages 1365--1375, Marseille, France. European
  Language Resources Association.

\bibitem[{Troiano et~al.(2019)Troiano, Pad{\'o}, and
  Klinger}]{troiano-etal-2019-crowdsourcing}
Enrica Troiano, Sebastian Pad{\'o}, and Roman Klinger. 2019.
\newblock \href {https://doi.org/10.18653/v1/P19-1391} {Crowdsourcing and
  validating event-focused emotion corpora for {G}erman and {E}nglish}.
\newblock In \emph{Proceedings of the 57th Annual Meeting of the {A}ssociation
  for {C}omputational {L}inguistics}, pages 4005--4011, Florence, Italy.
  Association for Computational Linguistics.

\bibitem[{Wemmer et~al.(2024)Wemmer, Labat, and
  Klinger}]{wemmer-etal-2024-emoprogress}
Eileen Wemmer, Sofie Labat, and Roman Klinger. 2024.
\newblock \href {https://aclanthology.org/2024.lrec-main.503/}
  {{E}mo{P}rogress: Cumulated emotion progression analysis in dreams and
  customer service dialogues}.
\newblock In \emph{Proceedings of the 2024 Joint International Conference on
  Computational Linguistics, Language Resources and Evaluation ({LREC}-{COLING}
  2024)}, pages 5660--5677, Torino, Italia. ELRA and ICCL.

\bibitem[{Widmann and Wich(2023)}]{widmann-wich-2023-creating}
Tobias Widmann and Maximilian Wich. 2023.
\newblock \href {https://doi.org/10.1017/pan.2022.15} {Creating and comparing
  dictionary, word embedding, and transformer-based models to measure discrete
  emotions in {G}erman political text}.
\newblock \emph{Political Analysis}, 31(4):626--641.

\bibitem[{Wu et~al.(2025)Wu, Cai, Liu, Zhu, Xue, Gong, Hirschberg, and
  Ma}]{wu-etal-2025-multimodal}
ChengYan Wu, Yiqiang Cai, Yang Liu, Pengxu Zhu, Yun Xue, Ziwei Gong, Julia
  Hirschberg, and Bolei Ma. 2025.
\newblock \href {https://doi.org/10.18653/v1/2025.findings-emnlp.332}
  {Multimodal emotion recognition in conversations: A survey of methods,
  trends, challenges and prospects}.
\newblock In \emph{Findings of the {A}ssociation for {C}omputational
  {L}inguistics: {EMNLP} 2025}, pages 6257--6274, Suzhou, China. Association
  for Computational Linguistics.

\bibitem[{Zahiri and Choi(2018)}]{zahiri-choi-2018-emotion}
Sayyed~M. Zahiri and Jinho~D. Choi. 2018.
\newblock \href {https://arxiv.org/abs/1708.04299} {Emotion detection on {TV}
  show transcripts with sequence-based convolutional neural networks}.
\newblock In \emph{The AAAI-18 Workshop on Affective Content Analysis}.

\bibitem[{Zhan et~al.(2023)Zhan, Ong, and Li}]{zhan-etal-2023-evaluating}
Hongli Zhan, Desmond~C. Ong, and Junyi~Jessy Li. 2023.
\newblock \href {https://doi.org/10.18653/v1/2023.findings-emnlp.962}
  {Evaluating subjective cognitive appraisals of emotions from large language
  models}.
\newblock In \emph{Findings of the {A}ssociation for {C}omputational
  {L}inguistics: {EMNLP} 2023}, pages 14418--14446, Singapore. Association for
  Computational Linguistics.

\bibitem[{Zhang et~al.(2024{\natexlab{a}})Zhang, Deng, Liu, Pan, and
  Bing}]{zhang-etal-2024-sentiment}
Wenxuan Zhang, Yue Deng, Bing Liu, Sinno Pan, and Lidong Bing.
  2024{\natexlab{a}}.
\newblock \href {https://doi.org/10.18653/v1/2024.findings-naacl.246}
  {Sentiment analysis in the era of large language models: A reality check}.
\newblock In \emph{Findings of the {A}ssociation for {C}omputational
  {L}inguistics: {NAACL} 2024}, pages 3881--3906, Mexico City, Mexico.
  Association for Computational Linguistics.

\bibitem[{Zhang et~al.(2024{\natexlab{b}})Zhang, Yang, Xu, Gao, Huang, Mu,
  Feng, Wang, Zhang, Song, and Yu}]{zhang-etal-2024-affective}
Yiqun Zhang, Xiaocui Yang, Xingle Xu, Zeran Gao, Yijie Huang, Shiyi Mu, Shi
  Feng, Daling Wang, Yifei Zhang, Kaisong Song, and Ge~Yu. 2024{\natexlab{b}}.
\newblock \href {https://arxiv.org/abs/2408.04638} {Affective computing in the
  era of large language models: A survey from the {NLP} perspective}.
\newblock \emph{Preprint}, arXiv:2408.04638.

\bibitem[{Zhao et~al.(2024)Zhao, Ren, Hessel, Cardie, Choi, and
  Deng}]{zhao-etal-2024-wildchat}
Wenting Zhao, Xiang Ren, Jack Hessel, Claire Cardie, Yejin Choi, and Yuntian
  Deng. 2024.
\newblock \href
  {https://proceedings.iclr.cc/paper_files/paper/2024/hash/9421261e06f1a63a352b068f1ac90609-Abstract-Conference.html}
  {{WildChat}: 1{M} {ChatGPT} interaction logs in the wild}.
\newblock In \emph{The Twelfth International Conference on Learning
  Representations}.

\bibitem[{Zheng et~al.(2024)Zheng, Chiang, Sheng, Li, Zhuang, Wu, Zhuang, Li,
  Lin, Xing, Gonzalez, Stoica, and Zhang}]{zheng-etal-2024-lmsys}
Lianmin Zheng, Wei-Lin Chiang, Ying Sheng, Tianle Li, Siyuan Zhuang, Zhanghao
  Wu, Yonghao Zhuang, Zhuohan Li, Zi~Lin, Eric~P. Xing, Joseph~E. Gonzalez, Ion
  Stoica, and Hao Zhang. 2024.
\newblock \href
  {https://proceedings.iclr.cc/paper_files/paper/2024/hash/5f9bfdfe3685e4ccdbc0e7fb29cccf2a-Abstract-Conference.html}
  {{LMSYS-Chat-1M}: A large-scale real-world {LLM} conversation dataset}.
\newblock In \emph{The Twelfth International Conference on Learning
  Representations}.

\end{thebibliography}
\end{document}